\def\ARXIVVERSION{}
\documentclass{article}
\ifdefined\ARXIVVERSION
\else
  \documentclass{article}
\fi
\newif\ifarxivversion
\ifdefined\ARXIVVERSION
  \arxivversiontrue
\else
  \arxivversionfalse
\fi

\usepackage{iclr2027_conference,times}
\ifarxivversion
  \iclrfinalcopy
\fi
\usepackage[utf8]{inputenc}
\usepackage[T1]{fontenc}
\usepackage{url}
\usepackage{hyperref}
\usepackage{booktabs}
\usepackage{amsmath,amssymb,amsthm}
\usepackage{graphicx}
\usepackage{xcolor}
\usepackage{enumitem}
\usepackage{algorithm}
\usepackage{algpseudocode}
\usepackage{caption}
\usepackage{float}

\graphicspath{{figures/}{./}}
\hypersetup{
  colorlinks=true,
  linkcolor=blue!55!black,
  citecolor=blue!55!black,
  urlcolor=blue!55!black,
}

\newcommand{\configs}{\mathcal{C}}
\newcommand{\dataset}{\mathcal{D}}
\newcommand{\pareto}{\mathcal{P}}
\newcommand{\ccgittins}{\textsc{CC-Gittins}}
\newtheorem{proposition}{Proposition}

\newcommand{\E}{\mathbb{E}}
\newcommand{\R}{\mathbb{R}}
\newcommand{\Normal}{\mathcal{N}}

\title{Which Model for Which Role?\\
Cost-Efficient Pareto Recommendation\\
for LLM Agentic Workflows}

\ifarxivversion
    \author{
    
    \setcounter{footnote}{0}
    \makebox[\dimexpr\textwidth-2\tabcolsep-3pt\relax][c]{%
        \bfseries
        Qian Xie$^{1,2}$\thanks{Equal contribution.}\hspace{0.5em}%
        \thanks{\hangindent=1.8em\hangafter=1 Corresponding authors:
        Qian Xie (qx66@cornell.edu);
        Wenyue Hua (wenyuehua@microsoft.com);
        Tianyi Peng (tianyi.peng@columbia.edu);
        Nairen Cao (caonairen@sufe.edu.cn).}%
        \qquad
        Wenyue Hua$^{3}$\footnotemark[1]\hspace{0.5em}\footnotemark[2]
        \qquad
        Yueli He$^{4}$
        \qquad
        Sripad Karne$^{1}$
        \qquad
        Armaan Agrawal$^{1}$
    }\\[6pt]
    \makebox[\dimexpr\textwidth-2\tabcolsep-3pt\relax][c]{%
        \bfseries
        Nikos Pagonas$^{1}$
        \qquad
        Eugene Wu$^{1}$
        \qquad
        Kostis Kaffes$^{1}$
        \qquad
        Tianyi Peng$^{1}$\footnotemark[2]
        \qquad
        Nairen Cao$^{5,4}$\footnotemark[2]
    }\\[10pt]
    \normalfont
    \makebox[\dimexpr\textwidth-2\tabcolsep-3pt\relax][c]{%
        $^{1}$Columbia University \quad
        $^{2}$Cornell University \quad
        $^{3}$Microsoft
    }\\
    \makebox[\dimexpr\textwidth-2\tabcolsep-3pt\relax][c]{%
        $^{4}$New York University \quad
        $^{5}$Shanghai University of Finance and Economics
    }
    }
\else
  \author{Anonymous Authors}
\fi

\ifarxivversion
  \makeatletter
  \let\input@path\@empty
  \makeatother
\fi

\begin{document}
\maketitle
\ifarxivversion
  \lhead{Preprint.}
\fi

\begin{abstract}
LLM agentic workflows increasingly employ different models for specialized roles. Users may prefer different trade-offs between task quality and deployment cost, sometimes also considering latency. Finding good assignments incurs a separate search cost from evaluating multiple candidates. We introduce \emph{Cost-Coupled Gittins} (\textsc{CC-Gittins}), a multi-objective extension of single-objective cost-aware Gittins search for Pareto recommendation. It exploits the coupling between evaluation and deployment costs: assignments that cost more to deploy also cost more to evaluate. The search alternates between seeking higher quality and lower deployment cost, reusing each evaluation to inform both searches. After each batch, \ccgittins{} recommends assignments that are nondominated under uncertainty-adjusted estimates of quality and deployment cost. We evaluate \textsc{CC-Gittins} on eight quality--cost matrices from workflows with two to five roles and up to 1,681 candidate assignments. We also evaluate the method with latency as a third objective. Across the eight benchmarks, \ccgittins{} recovers a substantial fraction of the true Pareto frontier at roughly 10\% of the exhaustive-evaluation cost, and recovers most of the frontier with few false positives by 30\%.
\ifarxivversion
\par\smallskip
\noindent\textbf{Code:}
\href{https://github.com/AgentOptimizer/agentopt-research}{\texttt{github.com/AgentOptimizer/agentopt-research}}
\fi
\end{abstract}

\section{Introduction}
\label{sec:introduction}

\begingroup
\setlength{\intextsep}{5pt}
\begin{figure}[H]
  \centering
  \captionsetup{font=small}
  \includegraphics[width=\linewidth,trim=8 14 8 14,clip]{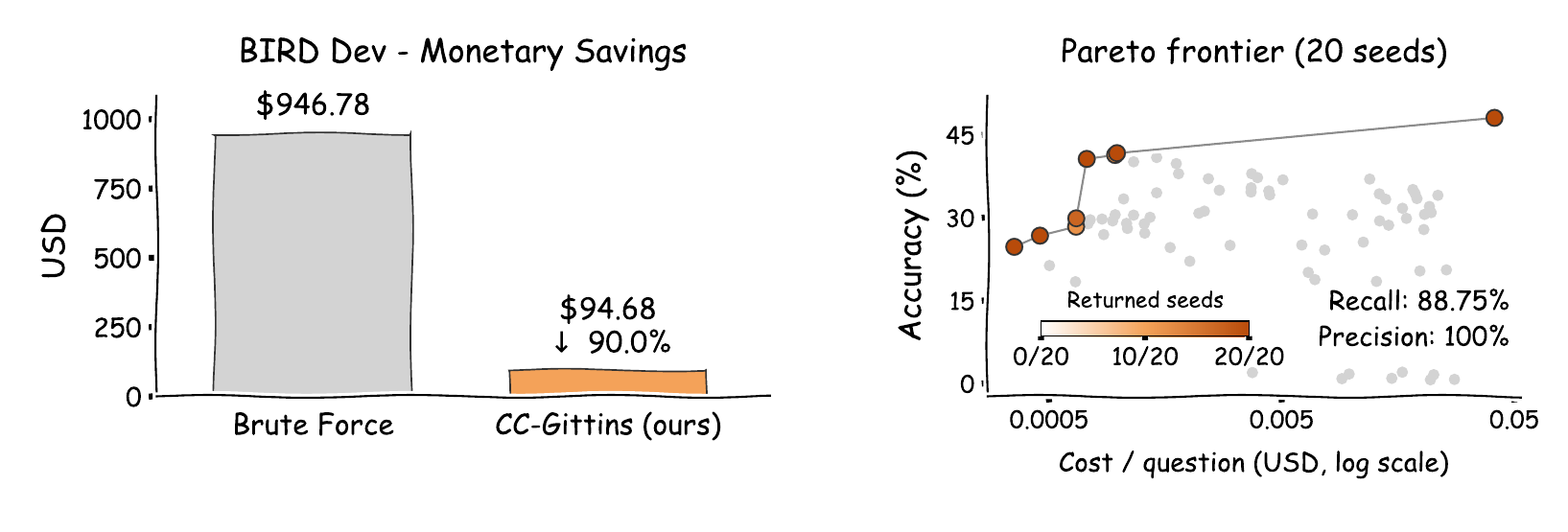}
  \caption{At 90\% lower evaluation cost than exhaustive evaluation,
  \ccgittins{} (ours) recovers most of the BIRD Dev Pareto frontier with no false
  positives (88.75\% mean recall and 100\% mean precision over 20 seeds).}
  \label{fig:bird-dev-cost-frontier}
\end{figure}
\endgroup

Large language model (LLM) agents increasingly combine multiple interacting
roles, such as planning, solving, searching, and verification
\citep{hua2026agentopt,kim2026teambench,jiang2026agentcard}.
These roles need not use the same underlying model, and using the most
capable models in every role can incur substantial deployment cost. This
creates a \emph{model assignment problem}---given a set of
candidate models for each role, which assignments best balance end-to-end task
quality, deployment cost, and, when response time matters, latency?
Since one role's output can affect the inputs and behavior of subsequent
roles, an assignment's quality generally cannot be inferred by evaluating
each model in isolation. The whole workflow must be treated as the
unit of evaluation.

Users may want to compare the task quality of inexpensive and more costly assignments before choosing what to deploy. A representative range of quality--cost trade-offs can support this decision without requiring every Pareto-optimal assignment to be identified.

Existing methods either report quality--cost frontiers over evaluated agent assignments (TeamBench, AgentCARD, and AgentOpt) \citep{kim2026teambench,jiang2026agentcard,hua2026agentopt}, or optimize a specified objective or constraint. In particular, AgentOpt reduces multi-objective search to single-objective optimization using a fixed weighted-sum utility \citep{hua2026agentopt}, SCOPE minimizes deployment cost subject to a quality threshold \citep{huang2026scope}, and VineLM adapts model choices at individual invocations under cost or latency budgets \citep{pagonas2026vinelm}. We study anytime recommendation of complete role assignments through costly evaluations, without fixing the preferred trade-off or final search cost in advance.


Exhaustive evaluation is costly: the number of assignments grows multiplicatively across roles, and each candidate must be run on multiple queries to estimate quality and deployment cost. Figure~\ref{fig:bird-dev-cost-frontier} illustrates quality--cost trade-offs recovered at a fraction of exhaustive evaluation cost.


A key observation is that \emph{deployment cost} and \emph{evaluation cost} are naturally coupled.  Let $D_i$ denote the mean deployment
cost per query for assignment $i$. Evaluating a batch of
$B$ comparable queries costs approximately
$\kappa_i \approx B D_i$. The total search cost accumulates
these charges across evaluated assignments, including those
never deployed. Consequently, configurations that are
expensive to deploy are also expensive to learn about.

\begin{figure}[t]
  \centering
  \includegraphics[width=0.98\linewidth]{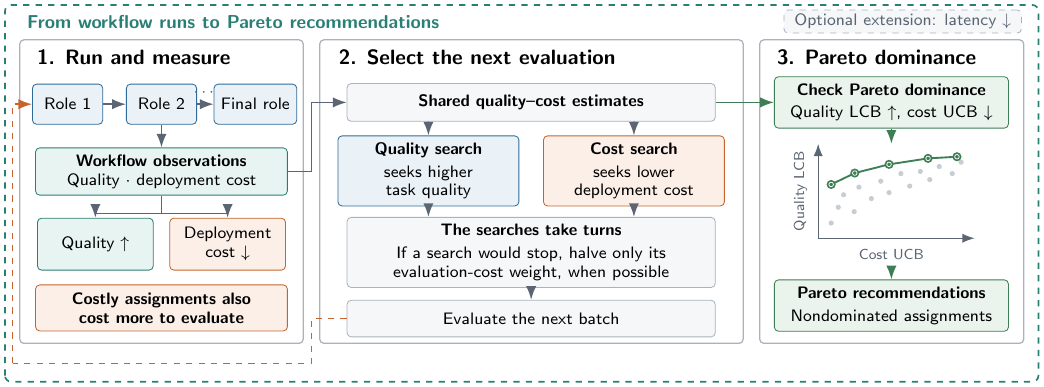}
\caption{Overview of \ccgittins{} in the quality--cost setting.
Workflow evaluations provide quality and deployment-cost observations.
Quality and cost controllers share observations, take turns choosing
evaluations, and independently relax their evaluation prices. After each
batch, Pareto dominance is checked using quality lower confidence bound
(LCB) and cost upper confidence bound (UCB) scores. Nondominated assignments
form the Pareto recommendations (green). When latency is included,
additional controllers share observations and recommendations use
three-objective Pareto dominance.}
  \label{fig:system-overview}
\end{figure}

We introduce \emph{Cost-Coupled Gittins} (\textsc{CC-Gittins}), a multi-objective method for anytime Pareto recommendation that exploits this coupling. For quality and cost, two search policies, called \emph{controllers}, take turns choosing evaluations: one seeks higher quality and the other lower deployment cost. The quality controller prices further evaluation using configuration-specific API costs. In the simplified model of Section~\ref{sec:cost-coupling-role}, its Gittins indices rank assignments by quality minus a deployment-cost penalty, so relaxing the evaluation-price multiplier can favor different quality--cost trade-offs. The controllers share observations. After each batch, assignments nondominated under uncertainty-adjusted quality and deployment-cost estimates form the recommendation. Figure~\ref{fig:system-overview} gives an overview.



We evaluate \ccgittins{} through offline replay on eight
quality--cost workflow benchmarks, including complete
HotpotQA and MathQA assignment matrices and six
SCOPE-derived instances. These benchmarks span two to five
roles and up to 1,681 candidate assignments. Compared with
five search baselines, \ccgittins{} achieves competitive
hypervolume regret and generational distance at low
search cost, although frontier coverage varies
across benchmarks. We further conduct controlled ablations
and evaluate the method with server-reported LLM latency as a third objective
on HotpotQA and MathQA.

\paragraph{Contributions.}
Our main contributions are as follows:
\begin{itemize}
    \item We formulate model assignment in multi-role LLM workflows as
    anytime recommendation of a useful range of Pareto trade-offs
    through costly end-to-end evaluations.

    \item We develop \ccgittins{}, a multi-objective extension of
    single-objective cost-aware Gittins search for anytime Pareto
    recommendation. Each batch updates the recommendation, so users can
    stop without fixing a final search budget or preferred trade-off.

    \item We implement a workflow evaluation harness\ifarxivversion{} in AgentOpt\fi{} and collect
    complete response matrices recording task quality, deployment cost, and
    LLM latency. These records support evaluation against exhaustive Pareto frontiers.

    \item Across eight quality--cost workflow benchmarks, \ccgittins{}
    recovers a substantial fraction of the true Pareto frontier at roughly
    $10\%$ of exhaustive evaluation cost, and most of the frontier with few
    false positives by $30\%$. We also evaluate the method with latency
    as a third objective.
\end{itemize}

\subsection{Related Work}
\label{sec:related}

\paragraph{Resource-aware LLM systems.}
Server-side systems improve scheduling, memory use, and serving throughput~\citep{kwon2023vllm,zheng2024sglang,luo2025autellix}. Client-side routing selects a model per request or call~\citep{chen2023frugalgpt,hu2024routerbench,ong2025routellm}. Agent configuration search instead assigns models to interacting roles and observes an end-to-end trajectory reward. TeamBench exhaustively studies the assignments of several providers to Planner, Executor, and Verifier~\citep{kim2026teambench}, while AgentCARD maps deployment-aware cost--accuracy frontiers over multi-role assignments~\citep{jiang2026agentcard}. These systems characterize configuration landscapes after evaluation.

\paragraph{Efficient LLM evaluation.}
BanditEval exploits low-rank response-matrix structure~\citep{zhou2025speeding},
SySRs uses paired comparisons to identify the best model~\citep{lyu2026cutting},
and PromptEval shares information across prompts and examples to estimate
multi-prompt performance~\citep{polo2024efficient}. Gittins indices have also
been adapted to account for evaluation costs in LLM model
selection~\citep{xie2026efficient}. These methods focus on scalar performance
estimation or single-objective model selection; \ccgittins{} extends
cost-aware Gittins search to multi-objective Pareto recommendation for
complete agent workflows.

\paragraph{Multi-objective LLM configuration optimization.}
SCOPE and VineLM consider multiple objectives through constrained
optimization: SCOPE minimizes deployment cost subject to a quality
threshold~\citep{huang2026scope}, while VineLM maximizes accuracy under cost
or latency budgets~\citep{pagonas2026vinelm}. Concurrent work by
\citet{xue2026cost} uses a frequentist approach: empirical gap elimination
for Pareto identification and hypervolume-per-cost UCB for online LLM
configuration selection. Its experiments vary models, prompts, and decoding settings for accuracy--latency trade-offs
under token-based evaluation budgets. We take a Bayesian approach to joint
model assignment across interacting agent roles, using Gittins indices to
guide cost-aware search while maintaining anytime quality--deployment-cost
recommendations. Our method exploits the coupling between search and
deployment costs.

\paragraph{Pareto-set search.}
General-purpose multi-objective bandits study Pareto-set identification under fixed-confidence or fixed-budget objectives~\citep{kone2023adaptive,kone2024bandit}. Hypervolume-based Bayesian optimization, including qEHVI and qNEHVI, selects evaluations by expected hypervolume improvement~\citep{daulton2020differentiable,daulton2021parallel}. These methods can be sample-efficient, but their standard acquisition criterion does not explicitly price the assignment-dependent API cost of each evaluation. \ccgittins{} includes that price in its scalar indices and updates its recommendation after each batch; we assess those recommendations at matched cumulative API spend.

\section{Pareto Recommendation and Evaluation}
\label{sec:problem}

\paragraph{Configurations and objectives.}
For a given workflow, we vary the model assigned to each role while keeping its role structure and orchestration rules fixed. The workflow has roles $1,\ldots,L$, with candidate model set $\mathcal M_\ell$ for role $\ell$. Let $\configs=\mathcal M_1\times\cdots\times\mathcal M_L$ be the set of complete model assignments and $K=|\configs|$ its size. We number these configurations $1,\ldots,K$; throughout, $i$ identifies one complete assignment. Configuration $i$ has task quality $Q_i$, to maximize, and mean deployment cost per query $D_i$, to minimize; Section~\ref{sec:additional-objectives} adds latency. Since role outputs affect subsequent execution, the full assignment is the unit of evaluation. The Pareto-optimal reference set is
\begin{equation}
 \pareto^\star=\{i:\nexists j\text{ with }Q_j\ge Q_i,\ D_j\le D_i
 \text{ and at least one strict inequality}\}.
\end{equation}
Our goal is an anytime recommendation set offering useful coverage of quality--cost trade-offs. We assess its proximity to this reference frontier and its coverage of trade-off volume and frontier regions. For offline replay, $Q_i,D_i$ are the exhaustive means on the fixed $N$-question benchmark. These averages describe the fixed benchmark; they do not guarantee performance on new queries or under a different query distribution.

\paragraph{Objective scaling and search cost.}
To compute the evaluation metrics, we map each objective to a fixed scale where larger values are better. The resulting objective vectors lie in $[0,1]^d$, with the origin as the hypervolume reference. In the quality--cost case, positive affine maps of $(Q_i,-D_i)$ give $y_i=(q_i,c_i)\in[0,1]^2$. Write $P^\star=\{y_i:i\in\pareto^\star\}$. Let $\widehat\pareto_t$ be the recommended configurations and $A_t=\{y_i:i\in\widehat\pareto_t\}$ their \emph{true} objective vectors. These vectors are available to the offline evaluator, not the selector. Total search cost $S_t=\sum_{s\le t}\kappa_s$ sums the batch charges $\kappa_s$ across all evaluated configurations. For exhaustive search cost $S_{\mathrm{full}}$, the recommendation snapshot at cumulative cost fraction $b$ is
\begin{equation}
 \widehat\pareto(b)=\widehat\pareto_{t(b)},\qquad
 t(b)=\max\{t:S_t\le bS_{\mathrm{full}}\}.
 \label{eq:anytime-budget}
\end{equation}
These cost fractions select retrospective reporting checkpoints and are not required inputs to \ccgittins{}. A snapshot is reported only if a recommendation is available at or below the corresponding cost threshold, including warm-start cost.

\paragraph{Evaluation metrics.}
For an offline evaluator, let $A$ be the nonempty set of recommended assignments at their true normalized coordinates, and let $d(x,S)$ be the Euclidean distance from $x$ to the nearest point in $S$. Hypervolume (HV) regret measures the volume dominated by the true frontier but not by the recommendation. Generational distance (GD) averages how far reported assignments lie from the true frontier. Inverted generational distance (IGD) averages how far true frontier points lie from the recommendation. We use
\begin{align}
 R_{\mathrm{HV}}(A)&=\operatorname{HV}(P^\star)-\operatorname{HV}(A),\\
 \operatorname{GD}(A,P^\star)&=|A|^{-1}\sum_{a\in A}d(a,P^\star),\\
 \operatorname{IGD}(A,P^\star)&=|P^\star|^{-1}\sum_{p^\star\in P^\star}d(p^\star,A).
 \label{eq:set-metrics}
\end{align}
These average or volume measures assess recommendations at a given search spend; they do not require exact frontier recovery or limit the number of returned assignments. They are not worst-case guarantees for individual assignments or frontier regions. All three metrics evaluate the returned recommendation at its true normalized coordinates, and GD and IGD use the complete set $A$.

\section{Cost-Coupled Gittins}
\label{sec:method}
Our goal is to progressively identify model assignments
with different quality--cost trade-offs while minimizing
evaluation costs. To achieve this, \ccgittins{} uses
two controllers, one focusing on quality and the other
on deployment cost. The controllers take turns selecting
assignments for evaluation and share their observations.
After each evaluation batch, we update the recommended
assignments based on Pareto dominance.
We next explain how the controllers select evaluations
and how the recommendations are constructed.

\subsection{How evaluation prices guide deployment trade-offs}
\label{sec:cost-coupling}

Deployment cost affects both the quality--cost trade-off
and the cost of evaluating an assignment. For comparable
benchmark and deployment queries, evaluating a batch of
$B$ queries for configuration $i$ costs approximately
$\kappa_i \approx B D_i$.
We use a positive estimate $\widehat{\kappa}_i$, fixed
after warm-up, to select evaluations, while actual API
charges contribute to the cumulative search cost $S_t$.
These costs may differ due to variations in query
difficulty, output length, and batch size.

\ccgittins{} incorporates these evaluation costs into
its Gittins indices. As a result, the quality controller
considers both potential quality improvements and the
cost of obtaining further observations. Since expensive
assignments also cost more to evaluate, this can favor
different quality--cost trade-offs. We characterize
this effect in Section~\ref{sec:cost-coupling-role}.



The quality controller uses task quality $Q_i$, while
the cost controller uses a transformed deployment cost
$C_i = 1-D_i/R$, where $R>0$ is fixed after warm-up.
We write
\begin{equation}
Y_i=(Q_i,C_i),\qquad
v_Q=(1,0),\quad v_C=(0,1).
\label{eq:axes}
\end{equation}
These coordinates are used for selecting evaluations
and differ from the normalized coordinates used for
offline evaluation in Section~\ref{sec:problem}.


\subsection{Shared Posterior and Gittins Indices}

Each evaluation batch provides observations of quality
and per-query deployment cost for the selected
configuration. Both controllers share these observations
and update the same posterior for each configuration.
For the unknown mean $\theta_{i,j}$ of
objective $j\in\{Q,D\}$, we use independent Gaussian
posterior marginals
\[
\theta_{i,j}\mid\dataset_{i,n}
\sim\Normal(m_{i,j,n},v_{i,j,n}),
\]
where $m_{i,j,n}$ and $v_{i,j,n}$ denote the posterior
mean and variance after $n$ observed queries.
The quality controller uses
\[
m^Q_{i,n}=m_{i,Q,n},\qquad
v^Q_{i,n}=v_{i,Q,n},
\]
while the cost controller uses
\[
m^C_{i,n}=1-m_{i,D,n}/R,\qquad
v^C_{i,n}=v_{i,D,n}/R^2.
\]
The coupling between evaluation and deployment costs
enters through evaluation prices, rather than posterior
correlations. Appendix~\ref{app:posterior} gives the
posterior calibration.

To select evaluations, each controller uses the
cost-aware Gittins index of
\citet{xie2026efficient}. For controller
$a\in\{Q,C\}$, let $k$ index the evaluation batches
of configuration $i$, and let $n_i(k)$ be its number
of observed queries at stage $k$. We use the shorthand
$m^a_{i,k}=m^a_{i,n_i(k)}$ and
$v^a_{i,k}=v^a_{i,n_i(k)}$.
Let $H_i$ denote the completion stage.

Given an outside option $\alpha$, define the centered
posterior mean $x=m^a_{i,k}-\alpha$.
For a positive per-batch evaluation price $\gamma$
and a mean-zero Gaussian increment $Z^a_{i,k}$,
the recursion is
\begin{align}
V_{i,H_i}^{a,\gamma}(x)
    &=\max\{0,x\},\\
W_{i,k}^{a,\gamma}(x)
    &=-\gamma+
      \E[V_{i,k+1}^{a,\gamma}(x+Z^a_{i,k})],
      \qquad
V_{i,k}^{a,\gamma}(x)
    =\max\{0,W_{i,k}^{a,\gamma}(x)\}.
\label{eq:scalar-dp}
\end{align}
The increment variance is the reduction in posterior
variance between consecutive stages. Evaluation may
stop at intermediate stages, while terminal utility
is obtained only upon completion.

Let $r^a_{i,k}(\gamma)$ be the zero of
$W_{i,k}^{a,\gamma}$. The two controllers use
\begin{equation}
G^Q_{i,k}
=m^Q_{i,k}-r^Q_{i,k}(\eta_Q\widehat\kappa_i),
\qquad
G^C_{i,k}
=m^C_{i,k}-r^C_{i,k}(\eta_C\widehat\kappa_i).
\label{eq:cc-index}
\end{equation}
The positive multipliers $\eta_Q$ and $\eta_C$,
initially one, control the evaluation-price penalties
and are adjusted independently as described next.
Both indices use one-dimensional dynamic programs.
Appendix~\ref{app:dp} gives their computation
and numerical details. The two-axis design also has computational advantages.
At matched evaluation counts, its mean policy runtime is lower
than that of the tested three- and five-controller variants on
all eight benchmarks
(Fig.~\ref{fig:ablation-directions-runtime},
Appendix~\ref{app:ablations}).

\subsection{Controller Scheduling and Asynchronous cost relaxation}
The cost ($C$) and quality ($Q$) controllers alternate turns
in the order $C,Q$. Each controller $a\in\{C,Q\}$ uses a
multiplier $\eta_a$, initially one, to scale the estimated
evaluation cost $\widehat\kappa_i$ in its Gittins index
(Eq.~\ref{eq:cc-index}). On its turn, the controller compares
the largest index among unfinished configurations with its
best completed terminal value $I_a$ (set $I_a=-\infty$ if
none is complete). If an unfinished configuration is preferred,
the controller evaluates one batch and updates the shared posteriors.

Otherwise, the controller would stop at its current
evaluation price. Instead, we halve only its multiplier
$\eta_a$ when possible, reducing the penalty for future
evaluations. The other controller's multiplier and actual
API charges remain unchanged. The search terminates when
both controllers reach stopping conditions that cannot be
relaxed without an intervening evaluation, an external
budget is reached, or all configurations are complete.
The round-robin schedule balances controller turns rather
than evaluation counts or API cost.
Appendix~\ref{app:dp} provides the numerical stopping rules.


\subsection{Why cost coupling exposes quality--cost trade-offs}
\label{sec:cost-coupling-role}

At a fixed uncertainty schedule and remaining horizon, raising the
information price lowers the scalar index (Appendix~\ref{app:proofs}).
Coupling makes that price depend on a deployment objective. To isolate its
role, consider a deterministic completion model: outcomes are known, but
terminal utility still requires paying for the remaining evaluation batches.

\begin{proposition}[Deterministic roles of the two exact axes]\label{prop:scalarization}
Suppose uncertainty has vanished, each unfinished configuration requires the same $h$ remaining full batches of size $B$ to collect terminal utility, and each batch costs exactly $\kappa_i=BD_i$. Among unfinished configurations,
\begin{align}
 G_i^Q(\eta_Q)&=Q_i-\lambda_QD_i, &\lambda_Q&=\eta_QhB,\nonumber\\
 G_i^C(\eta_C)&=1-\lambda_CD_i, &\lambda_C&=R^{-1}+\eta_ChB>0.
 \label{eq:implicit-scalarization}
\end{align}
Thus the quality axis ranks a quality--cost scalarization, while the cost axis ranks configurations by increasing deployment cost.
\end{proposition}

Coupled prices favor a more expensive assignment when
$\Delta Q>\lambda_Q\Delta D$; lowering $\eta_Q$ relaxes this requirement
at fixed $h$. This can expose \emph{supported} trade-offs, meaning frontier
points maximizing $Q_i-\lambda D_i$ for some $\lambda\ge0$, while the cost
axis targets the inexpensive endpoint. Equal evaluation prices would
preserve the quality ranking. Bayesian uncertainty, unequal remaining
horizons, and estimated prices break this exact scalarization
interpretation. Discrete price relaxation need not visit every supported
point, while unsupported points lie outside this linear argument. Thus
the policy has no complete-frontier recovery guarantee.

\subsection{Anytime conservative Pareto recommendation}
\label{sec:anytime-pareto}

Acquisition prices choose which evidence to buy; recommendations are formed
from all configurations, rather than only controller winners. For the fixed
$N$-question benchmark, let $S_{i,j,n}$ be the sum of $n$ observed outcomes
and $\sigma_j^2$ the frozen per-question residual variance. The predicted
mean and variance of the final benchmark mean are
\begin{equation}
 M_{i,j,n}=\frac{S_{i,j,n}+(N-n)m_{i,j,n}}{N},\qquad
 U_{i,j,n}=\frac{(N-n)^2v_{i,j,n}+(N-n)\sigma_j^2}{N^2}.
 \label{eq:finite-bridge}
\end{equation}
At completion, this mean is the observed benchmark mean and its variance is
zero. After only a few evaluations, however, an assignment may appear to
offer high quality or low cost while its full-benchmark means remain
uncertain. Following the scalar uncertainty-penalized recommendation design
of \citet{xie2026efficient}, we form quality-LCB and cost-UCB scores for
reporting:
\begin{equation}
 Q^-_{i,n}=M_{i,Q,n}-\beta\sqrt{U_{i,Q,n}},\qquad
 D^+_{i,n}=M_{i,D,n}+\beta\sqrt{U_{i,D,n}},\qquad \beta=1.
 \label{eq:pareto-recommendation}
\end{equation}
For a Pareto recommendation, these paired scores define one point per assignment. Using each configuration's current observation count $n_i$, configuration $j$ dominates $i$ in this score space if $Q^-_{j,n_j}\ge Q^-_{i,n_i}$ and $D^+_{j,n_j}\le D^+_{i,n_i}$, with at least one strict inequality. Return all configurations that are nondominated under this relation. Partially evaluated configurations remain eligible, and there is no restriction to one recommendation per controller. At exhaustive completion, the recommended set equals the true benchmark frontier. The current set is available after warm-up and after each subsequent batch, allowing users to stop without specifying a final evaluation budget. Figures~\ref{fig:estimated-actual-seed42-10} and~\ref{fig:estimated-actual-seed42-30} illustrate the recommended configurations and quality--cost estimates available to users, comparing \ccgittins{} with the five baselines.

This adjustment aims to reduce unstable off-front recommendations but is not a simultaneous or time-uniform confidence certificate. It does not guarantee that every reported assignment lies near the true frontier. The single-objective Gittins results do not establish Bayes optimality for the full multi-objective search method.

\subsection{Additional deployment objectives}
\label{sec:additional-objectives}

Since evaluation costs scale with deployment costs, the evaluation-cost penalty also makes controllers targeting other objectives sensitive to deployment cost. With $d$ objectives, let $z_i\in\R^{d-1}$ collect non-cost desirabilities. For a \emph{linear} utility $U_v(i)=v^\top z_i$, $v\in\Delta^{d-2}$, Proposition~\ref{prop:scalarization}'s deterministic assumptions give the ranking $v^\top z_i-\lambda D_i$; the explicit non-cost direction family has $d-2$ dimensions.

For quality $Q_i$, mean deployment cost $D_i$, and mean latency $L_i$, let $\ell_i=1-L_i/T$, where $T>0$ is a latency scale fixed after warm-up. Alongside the pure quality, latency, and cost controllers, our three-objective experiment adds a balanced quality--latency controller with terminal utility
\begin{equation}
 U_{\mathrm{bal}}(i)=\min\{Q_i,\ell_i\},\qquad
 G^{\mathrm{det}}_{i,\mathrm{bal}}=U_{\mathrm{bal}}(i)-\lambda D_i.
 \label{eq:three-objective-index}
\end{equation}
Taking the minimum favors assignments with both high quality and low latency. All four controllers share observations and independently relax their API evaluation-cost penalties. Algorithm~\ref{alg:cc} gives the common replay procedure; Appendix~\ref{app:extension} specifies the three-objective indices, recommendation rule, and metrics.

\section{Experiments}
\label{sec:experiments}

\subsection{Experimental setup}

\paragraph{Benchmarks.}
Our quality--cost evaluation uses eight exhaustive or replay-complete matrices across five workflow families. We evaluate $10^2=100$ assignments on all 200 questions of each of HotpotQA~\citep{yang2018hotpotqa} (planner and search-enabled solver) and MathQA~\citep{amini2019mathqa} (answerer and critic). Six further instances use SCOPE's released query-level evaluations~\citep{huang2026scope}, covering three-to-five-role text-to-SQL, data-transformation, and data-integration workflows. These configurations form subsets of the nominal assignment spaces, each evaluated on every benchmark question. All methods replay the same complete score--cost matrices (Appendix~\ref{app:benchmark-matrices}).

\paragraph{Evaluation protocol and metrics.}
The offline replay reveals only purchased observations. \ccgittins{} uses four warm questions per configuration, batches of up to four unseen questions, and two exact-axis controllers with independent price relaxation. Both quality and deployment-cost prior variances are estimated from warm-start observations. Comparisons use 20 matched question-order seeds (42--61) and charge observed API costs, including warm-up. Metric scales are fixed across methods and seeds. We report HV regret, GD, and IGD (Section~\ref{sec:problem}), dividing HV regret by $\operatorname{HV}(P^\star)$ for the quality--cost plots; lower is better.

\paragraph{Baselines.}
We compare \ccgittins{} with EGE-SH for fixed-budget Pareto-set identification~\citep{kone2024bandit}, the adaptive Pareto-exploration sampling rule of APE-$k$~\citep{kone2023adaptive}, and qNEHVI, a Bayesian optimization method that selects evaluations by noisy expected hypervolume improvement~\citep{daulton2021parallel}. We also compare \ccgittins{} with two random search baselines: \emph{random questions}, which evaluates all assignments on the same randomly drawn benchmark questions, and \emph{random configurations}, which fully evaluates randomly drawn assignments. Appendix~\ref{app:baselines} gives their replay protocols.

\subsection{Quality--Cost Pareto Recommendation}

\begin{figure}[t]
  \centering
  \includegraphics[width=0.98\linewidth]{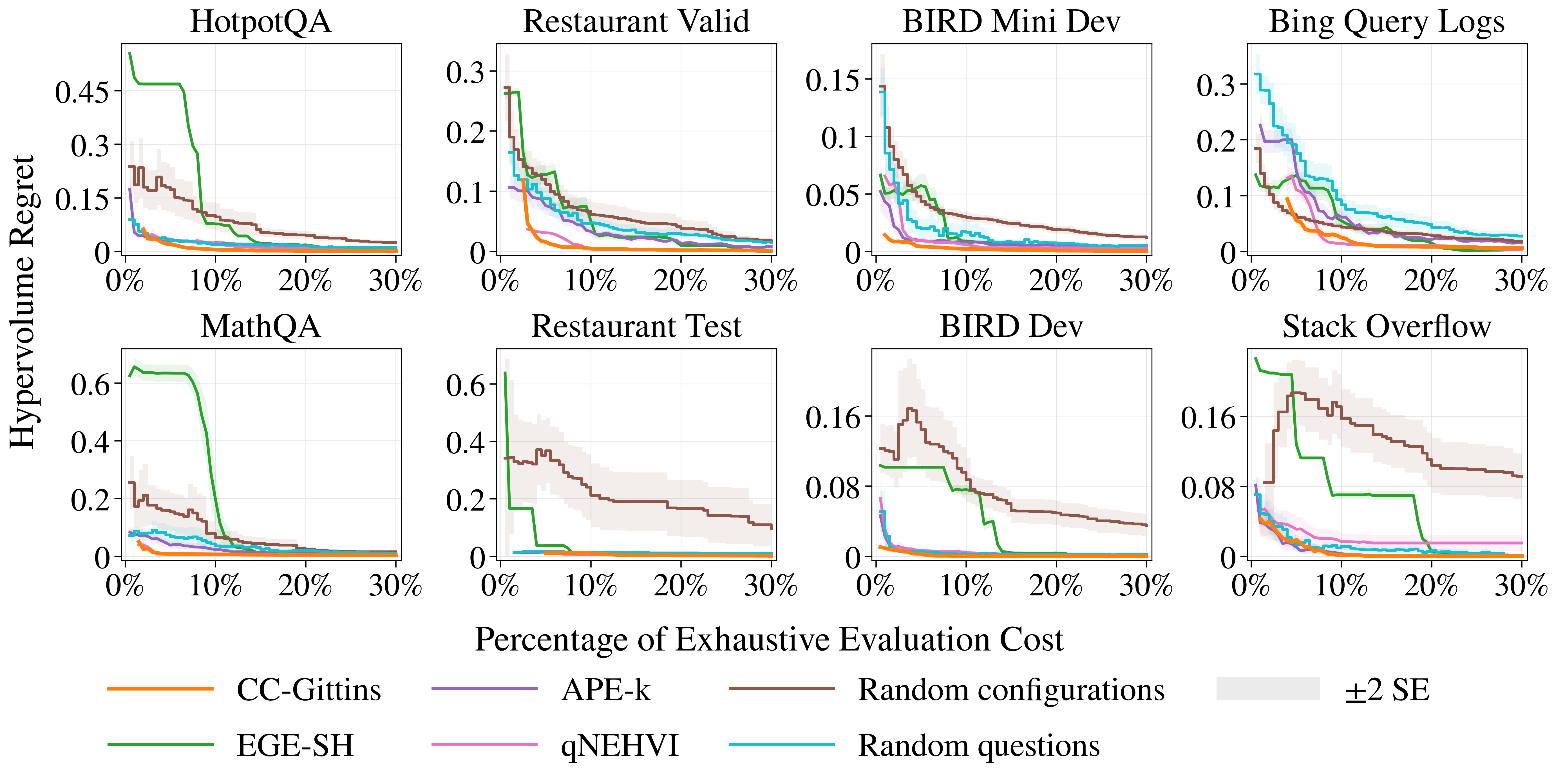}
  \caption{Hypervolume (HV) regret on eight benchmarks over $0$--$30\%$ of exhaustive evaluation cost (lower is better). \ccgittins{} is generally competitive with the baselines across benchmarks and the displayed cost range. All methods use 20 matched seeds; curves report seed means and bands show $\pm2$ standard errors.}
  \label{fig:anytime-hv}
\end{figure}

\begin{figure}[t]
  \centering
  \includegraphics[width=0.98\linewidth]{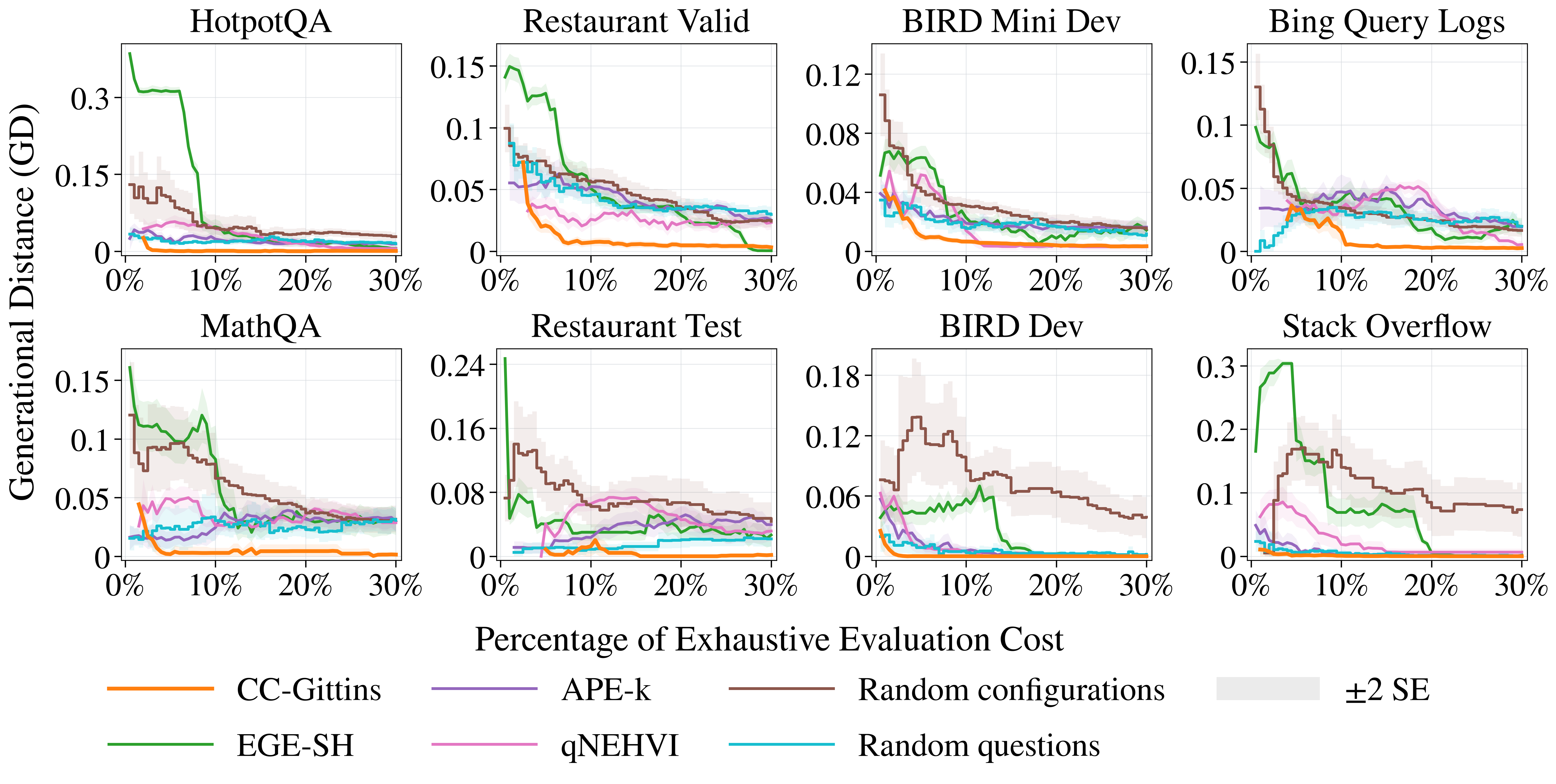}
  \caption{Generational distance (GD) on eight benchmarks over $0$--$30\%$ of exhaustive API evaluation cost (lower is better). \ccgittins{} is generally competitive with the baselines across benchmarks and the displayed cost range. Curves show seed means; bands show $\pm2$ standard errors.}
  \label{fig:anytime-gd}
\end{figure}

\paragraph{Main findings.}
At matched search costs up to $30\%$ of exhaustive evaluation, \ccgittins{} is generally competitive with the five baselines in HV regret and GD across eight benchmarks (Figs.~\ref{fig:anytime-hv}--\ref{fig:anytime-gd}). Partial evaluation can thus return assignments close to the exhaustive quality--cost frontier.

\begin{figure}[t]
  \centering
  \includegraphics[width=0.98\linewidth]{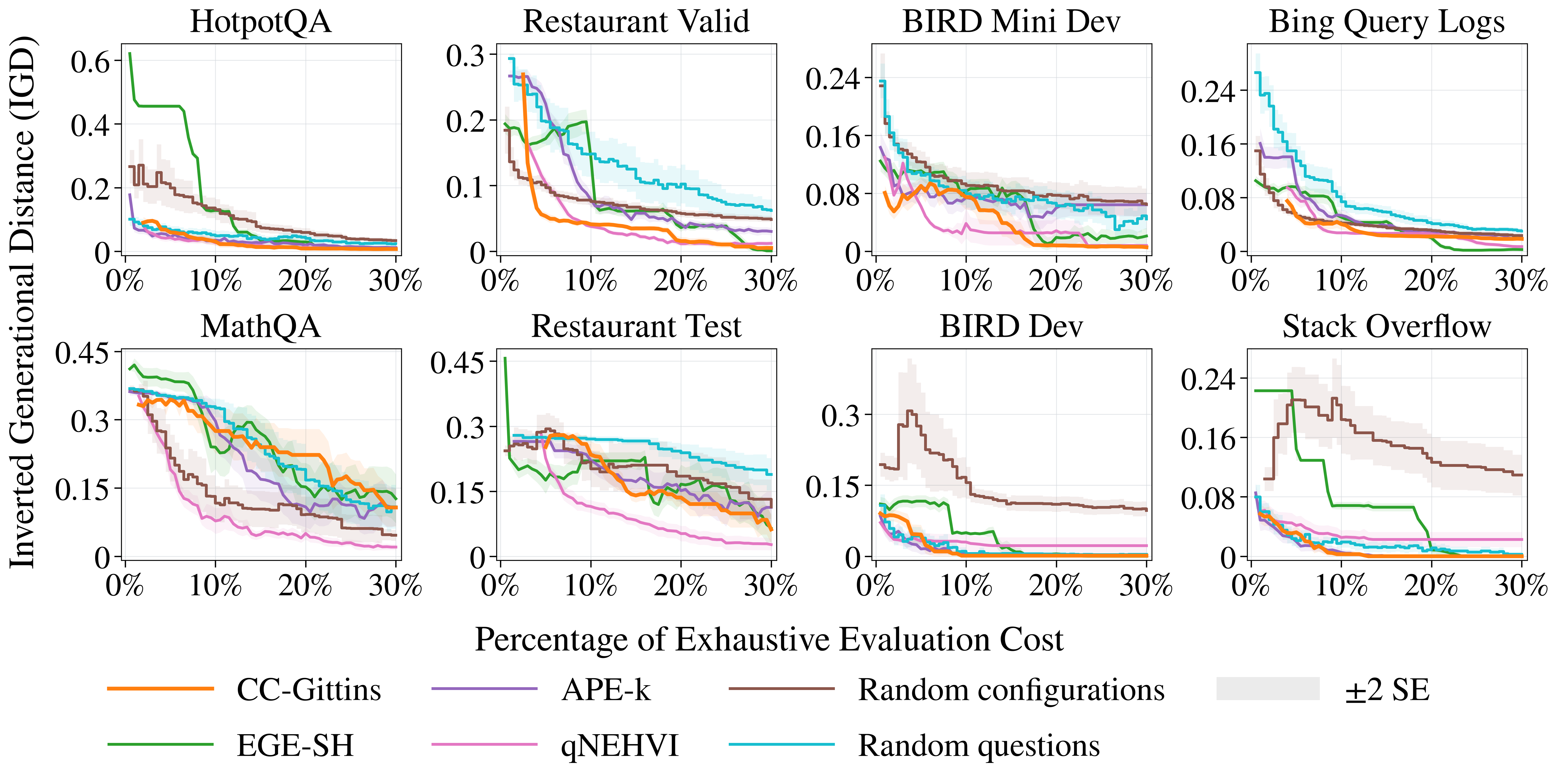}
  \caption{Inverted generational distance (IGD) on eight benchmarks over $0$--$30\%$ of exhaustive API evaluation cost (lower is better). \ccgittins{} achieves competitive frontier coverage, but does not consistently outperform the baselines. Curves show seed means; bands show $\pm2$ standard errors.}
  \label{fig:anytime-igd}
\end{figure}

\begin{figure}[t]
  \centering
  \includegraphics[width=0.98\linewidth]{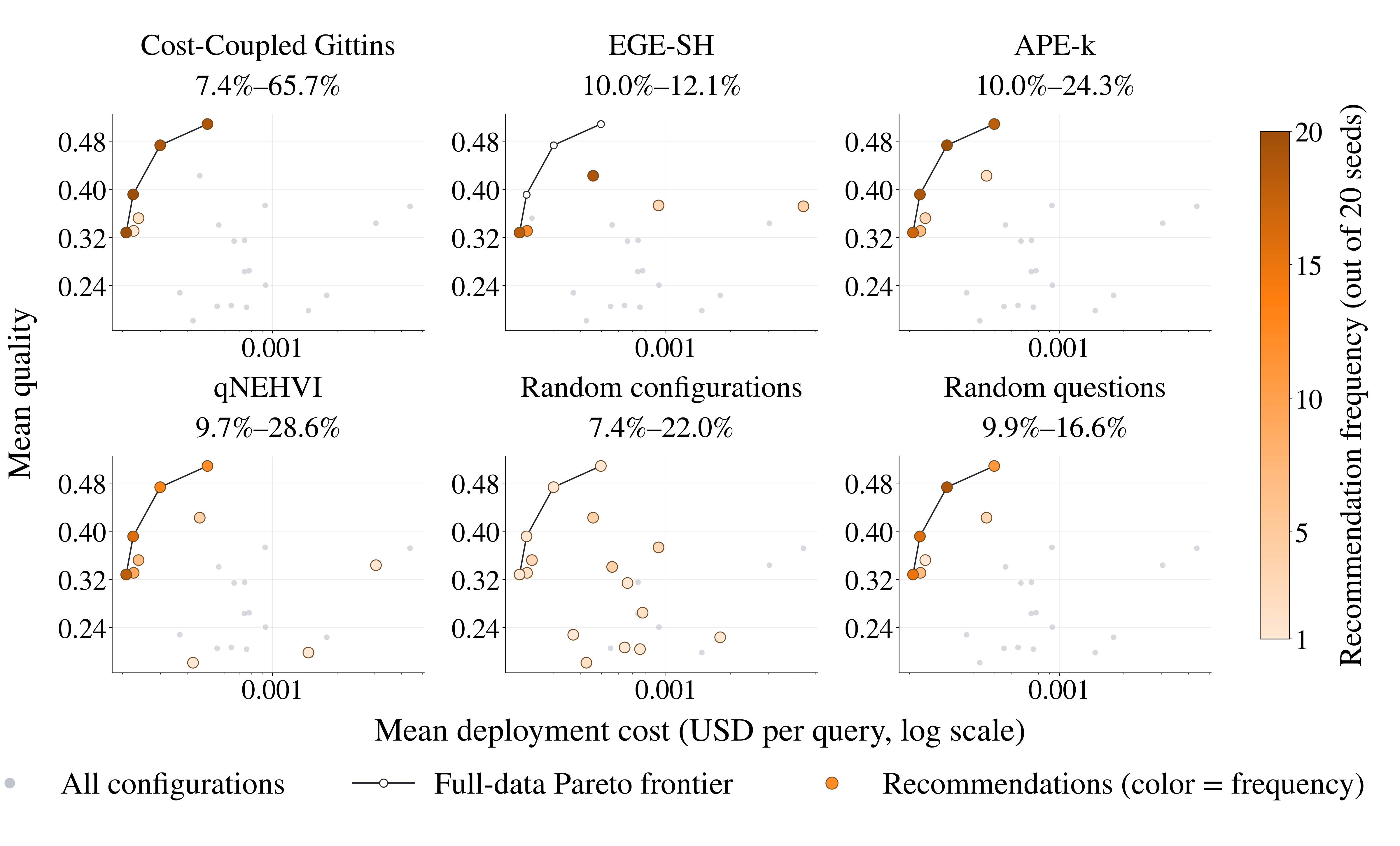}
  \caption{\ccgittins{} recovers nearly the entire Stack Overflow frontier at 10\% of exhaustive API evaluation cost, with few off-front recommendations (mean precision $0.97$, recall $0.98$). Panels compare \ccgittins{} with five baselines using exhaustive mean quality and deployment cost to locate configurations. Gray points show the full configuration set; black lines show the full-data Pareto frontier. Colored markers indicate recommendation frequency across 20 matched seeds. The line beneath each method name reports the mean start--end interval over which the checkpoint recommendation remains stable, as a fraction of exhaustive search cost.}
  \label{fig:stackoverflow-methods-10}
\end{figure}

\paragraph{Recommendation example.}
On Stack Overflow, \ccgittins{} nearly recovers the entire frontier at $10\%$ of exhaustive evaluation cost (mean precision $0.97$, recall $0.98$); both reach $1.00$ at $30\%$ (Fig.~\ref{fig:stackoverflow-methods-10}). Two acquisition controllers can return many trade-offs because the shared recommendation rule considers all configurations nondominated under uncertainty-adjusted estimates.

\paragraph{Coverage limitations.}
On HotpotQA, mean recall is $0.58$ at $10\%$ cost and $0.74$ at $30\%$, despite precision of $0.97$ and $0.99$, respectively. We prioritize representative Pareto-optimal recommendations: uncertainty penalties aim to reduce off-front selections but may exclude uncertain true Pareto points. This is consistent with low GD but weaker IGD (Fig.~\ref{fig:anytime-igd}); low HV regret can persist when retained points cover most dominated volume. On MathQA at $30\%$ cost, random configuration search achieves lower IGD but higher HV regret and GD than \ccgittins{}.

\paragraph{Ablation findings.}
At the $30\%$ cost checkpoint, combining configuration-specific evaluation prices with quality and cost controllers yields the lowest average HV regret, GD, and IGD across eight benchmarks among the cost-mechanism controls. Additional controllers improve particular metrics but do not uniformly outperform the two-axis design (Appendix~\ref{app:ablations}).

\paragraph{Weighted-score coverage.}
Across all eight benchmark matrices, the \emph{supported subset}
(Section~\ref{sec:cost-coupling-role}) loses less than $0.10\%$ of full-frontier
hypervolume despite omitting some Pareto points. This full-information oracle considers all weights, showing that weighted
scores can represent most dominated area without guaranteeing
\ccgittins{}'s performance under limited search cost and discrete price
relaxation.
Appendix~\ref{app:supported-front-diagnostic} details the geometry and
representative results.

\subsection{Latency-aware three-objective evaluation}
\label{sec:three-objective-results}

We evaluate the four-controller policy
(Section~\ref{sec:additional-objectives}), with direction-local $\eta$
decay, on the HotpotQA and MathQA matrices across 20 matched seeds,
comparing with the five baselines at matched search costs
(Appendix~\ref{app:extension}).

\ifarxivversion
\begin{figure}[!ht]
\else
\begin{figure}[t]
\fi
  \centering
  \includegraphics[width=0.98\linewidth]{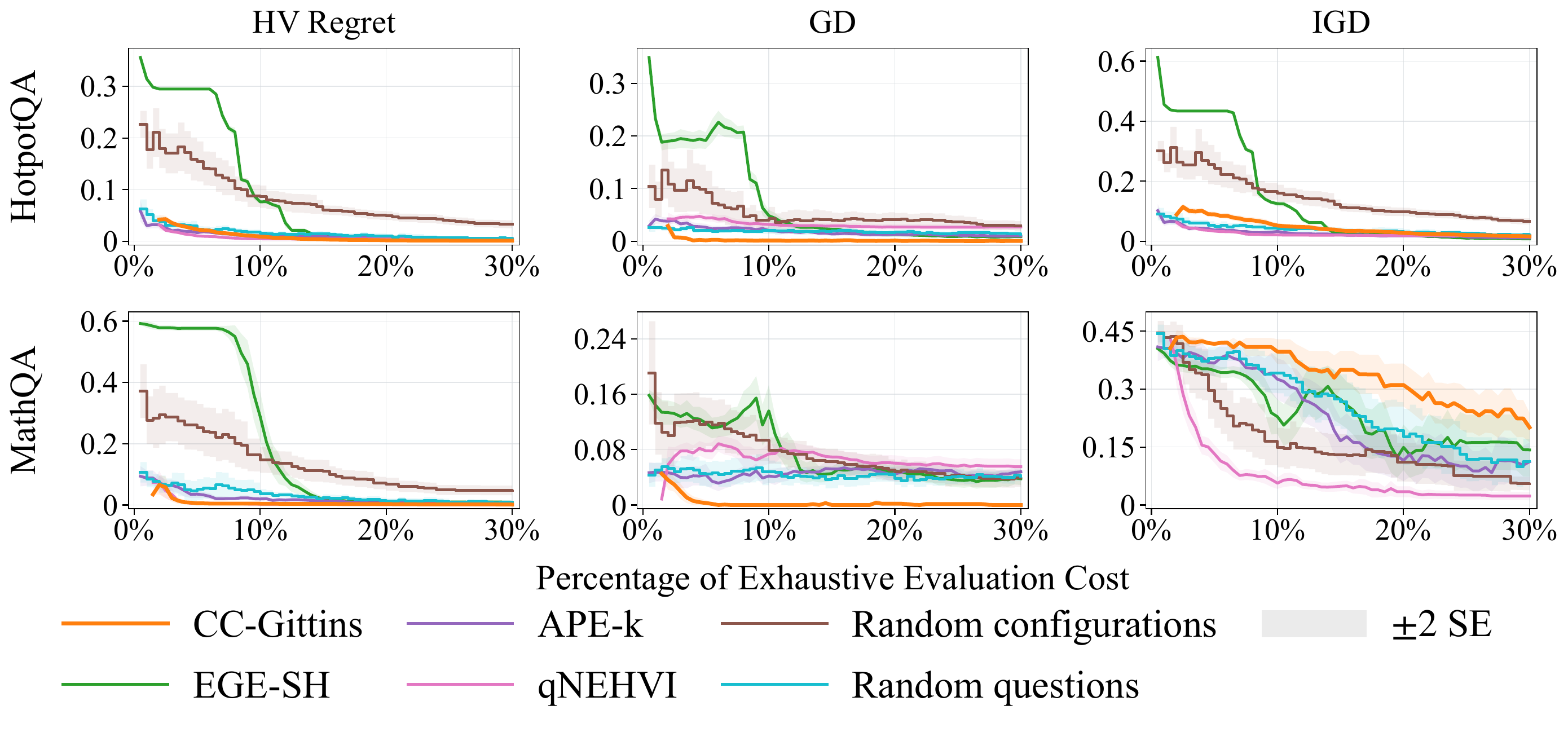}
  \caption{Three-objective evaluation of quality, deployment cost, and cumulative server-reported LLM latency on HotpotQA (top) and MathQA (bottom) over 0--30\% of exhaustive API evaluation cost. \ccgittins{} uses four controllers. Curves show seed means; bands show $\pm2$ standard errors.}
  \label{fig:three-objective-four}
\end{figure}

At $30\%$ of exhaustive evaluation cost, \ccgittins{} usually outperforms
the baselines in mean HV regret and GD, while its IGD is less competitive,
particularly on MathQA. Pareto precision is high ($99.6\%$ and $100.0\%$), while frontier
recovery remains incomplete (recall: $80.3\%$ and $54.1\%$, respectively).
Appendix~\ref{app:three-objective-ablations} assesses acquisition-cost
pricing, controller directions, direction-local continuation, and
warm-start prior calibration.

\section{Discussion and Conclusions}
\label{sec:conclusions-future-work}

\ccgittins{} exploits the coupling between deployment and evaluation costs for anytime multi-objective Pareto recommendation. Across eight quality--cost benchmarks, it recovers substantial portions of the exhaustive Pareto frontier at low evaluation cost. Experiments including latency also demonstrate applicability to three-objective Pareto recommendation.

We consider static role-level assignments under fixed orchestration. Complete frontier recovery is not guaranteed, and controller design remains empirical. Future work could adapt model choices, retries, and execution paths to intermediate outcomes.

\subsection*{Reproducibility statement}
Experiments use offline replay of fixed benchmark matrices without new model API calls. Appendices~\ref{app:dp}--\ref{app:implementation} document posterior calibration, index computation and stopping, the three-objective extension, and data sources, pricing, settings, and baselines. Plotting and ablation protocols are in Appendices~\ref{app:plot-protocol}--\ref{app:three-objective-ablations}.
\ifarxivversion
Code, benchmark matrices, and replay instructions are available at \url{https://github.com/AgentOptimizer/agentopt-research}.
\else
Code and a video presenting the experimental results are provided in the supplementary material.
\fi

\subsection*{AI use statement}
The authors used generative AI tools, including LLM-based assistants, to draft and revise parts of the manuscript, improve the clarity of the writing, and assist with \LaTeX{} formatting. The tools also assisted in developing mathematical proofs by working out detailed derivations and proof steps from statements of theoretical results, proof sketches, and arguments provided by the authors. The authors checked cited sources, mathematical arguments, and AI-assisted text, revised them as needed, and take full responsibility for the paper's contents.

Separately, LLMs are components of the agent configurations evaluated in this work. The quality--cost experiments replay HotpotQA and MathQA response matrices collected with our workflow evaluation harness and six SCOPE-derived instances; the three-objective replay uses the same two matrices with recorded latency. The replay makes no new model API calls. Appendix~\ref{app:benchmark-matrices} documents data sources and processing.

\ifarxivversion
\section*{Acknowledgements}
QX was supported in part by the Columbia AI Agents Initiative at Columbia Business School.
This research received funding from DAPLab corporate support in the form of funding and/or compute from Amazon, IntellectAI, Infosys, Tidalwave, Veris, Shopify, Microsoft, Thinking Machines, Dandy, Perplexity, and Daytona.
The authors also thank Wenxin Zhang for a helpful discussion and for suggestions on an earlier draft of the manuscript.
\fi

\clearpage
\IfFileExists{iclr_submission/references.bib}{%
  \bibliographystyle{iclr_submission/iclr2027_conference}
  \bibliography{iclr_submission/references}
}{%
  \bibliographystyle{iclr2027_conference}
  \bibliography{references}

\begin{thebibliography}{22}
\providecommand{\natexlab}[1]{#1}
\providecommand{\url}[1]{\texttt{#1}}
\expandafter\ifx\csname urlstyle\endcsname\relax
  \providecommand{\doi}[1]{doi: #1}\else
  \providecommand{\doi}{doi: \begingroup \urlstyle{rm}\Url}\fi

\bibitem[Amini et~al.(2019)Amini, Gabriel, Lin, Koncel-Kedziorski, Choi, and
  Hajishirzi]{amini2019mathqa}
Aida Amini, Saadia Gabriel, Shanchuan Lin, Rik Koncel-Kedziorski, Yejin Choi,
  and Hannaneh Hajishirzi.
\newblock {MathQA}: Towards interpretable math word problem solving with
  operation-based formalisms.
\newblock In \emph{Proceedings of NAACL-HLT}, pp.\  2357--2367, 2019.
\newblock \doi{10.18653/v1/N19-1245}.

\bibitem[Chen et~al.(2023)Chen, Zaharia, and Zou]{chen2023frugalgpt}
Lingjiao Chen, Matei Zaharia, and James Zou.
\newblock {FrugalGPT}: How to use large language models while reducing cost and
  improving performance.
\newblock \emph{arXiv preprint arXiv:2305.05176}, 2023.

\bibitem[Daulton et~al.(2020)Daulton, Balandat, and
  Bakshy]{daulton2020differentiable}
Samuel Daulton, Maximilian Balandat, and Eytan Bakshy.
\newblock Differentiable expected hypervolume improvement for parallel
  multi-objective {Bayesian} optimization.
\newblock In \emph{Advances in Neural Information Processing Systems},
  volume~33, pp.\  9851--9864, 2020.
\newblock URL
  \url{https://papers.neurips.cc/paper_files/paper/2020/hash/6fec24eac8f18ed793f5eaad3dd7977c-Abstract.html}.

\bibitem[Daulton et~al.(2021)Daulton, Balandat, and
  Bakshy]{daulton2021parallel}
Samuel Daulton, Maximilian Balandat, and Eytan Bakshy.
\newblock Parallel {Bayesian} optimization of multiple noisy objectives with
  expected hypervolume improvement.
\newblock In \emph{Advances in Neural Information Processing Systems},
  volume~34, pp.\  2187--2200, 2021.
\newblock URL
  \url{https://proceedings.neurips.cc/paper/2021/hash/11704817e347269b7254e744b5e22dac-Abstract.html}.

\bibitem[Hu et~al.(2024)Hu, Bieker, Li, Jiang, Keigwin, Ranganath, Keutzer, and
  Upadhyay]{hu2024routerbench}
Qitian~Jason Hu, Jacob Bieker, Xiuyu Li, Nan Jiang, Benjamin Keigwin, Gaurav
  Ranganath, Kurt Keutzer, and Shriyash~Kaustubh Upadhyay.
\newblock {RouterBench}: A benchmark for multi-{LLM} routing system.
\newblock \emph{arXiv preprint arXiv:2403.12031}, 2024.

\bibitem[Hua et~al.(2026)Hua, Karne, Xie, Agrawal, Pagonas, Kaffes, and
  Peng]{hua2026agentopt}
Wenyue Hua, Sripad Karne, Qian Xie, Armaan Agrawal, Nikos Pagonas, Kostis
  Kaffes, and Tianyi Peng.
\newblock {AgentOpt} v0.1 technical report: Client-side optimization for
  {LLM}-based agent.
\newblock \emph{arXiv preprint arXiv:2604.06296}, 2026.

\bibitem[Huang et~al.(2026)Huang, Zhang, Jin, and Xiao]{huang2026scope}
Yiqian Huang, Shiqi Zhang, Tianyuan Jin, and Xiaokui Xiao.
\newblock {SCOPE}: Cost-efficient model selection for compound {AI} systems
  under quality constraints.
\newblock In \emph{Proceedings of the 32nd ACM SIGKDD Conference on Knowledge
  Discovery and Data Mining V. 2}, pp.\  1922--1931. Association for Computing
  Machinery, 2026.
\newblock \doi{10.1145/3770855.3818067}.

\bibitem[Jiang et~al.(2026)Jiang, Cheng, Huang, Zhao, Lu, Dong, Li, Ponti, and
  Mai]{jiang2026agentcard}
Yinsicheng Jiang, Liang Cheng, Yeqi Huang, Yufan Zhao, Zhan Lu, Li~Dong, Wenda
  Li, Edoardo Ponti, and Luo Mai.
\newblock Specialize roles, mix deployments: Pushing the cost--accuracy
  frontier of {LLM} agent teams.
\newblock \emph{arXiv preprint arXiv:2606.20629}, 2026.

\bibitem[Kim et~al.(2026)Kim, Park, Kim, Park, Schmidgall, Rahman, Park,
  Breazeal, Liu, Palangi, Park, and McDuff]{kim2026teambench}
Yubin Kim, Chanwoo Park, Taehan Kim, Eugene Park, Samuel Schmidgall, Salman
  Rahman, Chunjong Park, Cynthia Breazeal, Xin Liu, Hamid Palangi, Hae~Won
  Park, and Daniel McDuff.
\newblock {TeamBench}: Evaluating agent coordination under enforced role
  separation.
\newblock \emph{arXiv preprint arXiv:2605.07073}, 2026.

\bibitem[Kone et~al.(2023)Kone, Kaufmann, and Richert]{kone2023adaptive}
Cyrille Kone, Emilie Kaufmann, and Laura Richert.
\newblock Adaptive algorithms for relaxed {Pareto} set identification.
\newblock In \emph{Advances in Neural Information Processing Systems},
  volume~36, pp.\  35190--35201, 2023.
\newblock \doi{10.52202/075280-1528}.
\newblock URL
  \url{https://proceedings.neurips.cc/paper_files/paper/2023/hash/6e976e7930460b5c3167a104ba8cc39c-Abstract-Conference.html}.

\bibitem[Kone et~al.(2024)Kone, Kaufmann, and Richert]{kone2024bandit}
Cyrille Kone, Emilie Kaufmann, and Laura Richert.
\newblock Bandit {Pareto} set identification: The fixed budget setting.
\newblock In \emph{Proceedings of the 27th International Conference on
  Artificial Intelligence and Statistics}, volume 238 of \emph{Proceedings of
  Machine Learning Research}, pp.\  2548--2556. PMLR, 2024.
\newblock URL \url{https://proceedings.mlr.press/v238/kone24a.html}.

\bibitem[Kwon et~al.(2023)Kwon, Li, Zhuang, Sheng, Zheng, Yu, Gonzalez, Zhang,
  and Stoica]{kwon2023vllm}
Woosuk Kwon, Zhuohan Li, Siyuan Zhuang, Ying Sheng, Lianmin Zheng, Cody~Hao Yu,
  Joseph~E. Gonzalez, Hao Zhang, and Ion Stoica.
\newblock Efficient memory management for large language model serving with
  {PagedAttention}.
\newblock In \emph{Proceedings of the 29th Symposium on Operating Systems
  Principles}, pp.\  611--626, 2023.

\bibitem[Luo et~al.(2025)Luo, Shi, Cai, Zhang, Wong, Wang, Wang, Huang, Chen,
  Gonzalez, et~al.]{luo2025autellix}
Michael Luo, Xiaoxiang Shi, Colin Cai, Tianjun Zhang, Justin Wong, Yichuan
  Wang, Chi Wang, Yanping Huang, Zhifeng Chen, Joseph~E. Gonzalez, et~al.
\newblock {Autellix}: An efficient serving engine for {LLM} agents as general
  programs.
\newblock \emph{arXiv preprint arXiv:2502.13965}, 2025.

\bibitem[Lyu et~al.(2026)Lyu, Nejma, Wegel, Yang, and Dorner]{lyu2026cutting}
Zifan Lyu, Chahine Nejma, Tobias Wegel, Fanny Yang, and Florian~E. Dorner.
\newblock Cutting {LLM} evaluation costs with {SySRs}: A bandit algorithm that
  provably exploits model similarity.
\newblock In \emph{International Conference on Machine Learning}, 2026.

\bibitem[Ong et~al.(2025)Ong, Almahairi, Wu, Chiang, Wu, Gonzalez, Kadous, and
  Stoica]{ong2025routellm}
Isaac Ong, Amjad Almahairi, Vincent Wu, Wei-Lin Chiang, Tianhao Wu, Joseph~E.
  Gonzalez, M~Waleed Kadous, and Ion Stoica.
\newblock {RouteLLM}: Learning to route {LLM}s from preference data.
\newblock In \emph{International Conference on Learning Representations},
  volume 2025, pp.\  34433--34448, 2025.

\bibitem[Pagonas et~al.(2026)Pagonas, Lou, Peng, Rubenstein, and
  Kaffes]{pagonas2026vinelm}
Nikos Pagonas, Matthew Lou, Tianyi Peng, Dan Rubenstein, and Kostis Kaffes.
\newblock {VineLM}: Trie-based fine-grained control for agentic workflows.
\newblock \emph{arXiv preprint arXiv:2605.23914}, 2026.

\bibitem[Polo et~al.(2024)Polo, Xu, Weber, Silva, Bhardwaj, Choshen,
  de~Oliveira, Sun, and Yurochkin]{polo2024efficient}
Felipe~Maia Polo, Ronald Xu, Lucas Weber, M{\'\i}rian Silva, Onkar Bhardwaj,
  Leshem Choshen, Allysson Flavio~Melo de~Oliveira, Yuekai Sun, and Mikhail
  Yurochkin.
\newblock Efficient multi-prompt evaluation of {LLM}s.
\newblock In \emph{Advances in Neural Information Processing Systems},
  volume~37, pp.\  22483--22512, 2024.
\newblock \doi{10.52202/079017-0707}.

\bibitem[Xie et~al.(2026)Xie, He, and Cao]{xie2026efficient}
Qian Xie, Yueli He, and Nairen Cao.
\newblock Efficient cost-aware {LLM} evaluation via {Bayesian} bandit {Gittins}
  indices.
\newblock In \emph{ICML 2026 Workshop on Decision-Making from Offline Datasets
  to Online Adaptation: Black-Box Optimization to Reinforcement Learning},
  2026.
\newblock URL \url{https://arxiv.org/abs/2609.25645}.

\bibitem[Xue et~al.(2026)Xue, Hong, Li, Wan, Cheng, and Qiu]{xue2026cost}
Bo~Xue, Zhi Hong, Jiayi Li, Yuanyu Wan, Ji~Cheng, and Shuang Qiu.
\newblock Cost-aware multi-objective bandits: Theory and application to
  budgeted {LLM} configuration evaluation.
\newblock \emph{arXiv preprint arXiv:2608.04333}, 2026.

\bibitem[Yang et~al.(2018)Yang, Qi, Zhang, Bengio, Cohen, Salakhutdinov, and
  Manning]{yang2018hotpotqa}
Zhilin Yang, Peng Qi, Saizheng Zhang, Yoshua Bengio, William Cohen, Ruslan
  Salakhutdinov, and Christopher~D. Manning.
\newblock {HotpotQA}: A dataset for diverse, explainable multi-hop question
  answering.
\newblock In \emph{Proceedings of the 2018 Conference on Empirical Methods in
  Natural Language Processing}, pp.\  2369--2380, 2018.
\newblock \doi{10.18653/v1/D18-1259}.

\bibitem[Zheng et~al.(2024)Zheng, Yin, Xie, Sun, Huang, Yu, Cao, Kozyrakis,
  Stoica, Gonzalez, et~al.]{zheng2024sglang}
Lianmin Zheng, Liangsheng Yin, Zhiqiang Xie, Chuyue Sun, Jeff Huang, Cody~Hao
  Yu, Shiyi Cao, Christos Kozyrakis, Ion Stoica, Joseph~E. Gonzalez, et~al.
\newblock {SGLang}: Efficient execution of structured language model programs.
\newblock \emph{Advances in Neural Information Processing Systems},
  37:\penalty0 62557--62583, 2024.
\newblock \doi{10.52202/079017-2000}.

\bibitem[Zhou et~al.(2025)Zhou, Belardi, Wu, Zhang, Gomes, Sun, and
  Weinberger]{zhou2025speeding}
Jin~Peng Zhou, Christian~K. Belardi, Ruihan Wu, Travis Zhang, Carla~P. Gomes,
  Wen Sun, and Kilian~Q. Weinberger.
\newblock On speeding up language model evaluation.
\newblock In \emph{International Conference on Learning Representations
  (ICLR)}, 2025.

\end{thebibliography}
}
\clearpage
\appendix

\section{Effects of Evaluation Costs on Gittins Indices}
\label{app:proofs}

\paragraph{Price monotonicity of the scalar recursion.}
Fix the transition schedule and terminal stage, and take $0<\gamma_1\le\gamma_2$. At the terminal stage $V_H^{\gamma_1}=V_H^{\gamma_2}=x_+$. If $V_{k+1}^{\gamma_1}\ge V_{k+1}^{\gamma_2}$ pointwise, then
\begin{equation}
 W_k^{\gamma_1}(x)=-\gamma_1+\E[V_{k+1}^{\gamma_1}(x+Z_k)]
 \ge-\gamma_2+\E[V_{k+1}^{\gamma_2}(x+Z_k)]=W_k^{\gamma_2}(x).
\end{equation}
Taking the maximum with zero preserves the inequality and proves it by backward induction. Continuation values are nondecreasing in $x$; consequently the nonnegative-continuation set for the higher price is contained in that for the lower price, and its threshold satisfies $r_k(\gamma_1)\le r_k(\gamma_2)$. With identical uncertainty schedules and remaining horizons, $m_i^Q\ge m_j^Q$ and $\widehat\kappa_i\le\widehat\kappa_j$ give
\begin{equation}
 G_i^Q-G_j^Q=(m_i^Q-m_j^Q)+r_k(\eta\widehat\kappa_j)-r_k(\eta\widehat\kappa_i)\ge0.
\end{equation}
We restrict to positive prices. At exactly zero cost, a nondegenerate Gaussian transition can make continuation strictly positive at every finite $x$, so a finite stopping root need not exist. This also motivates the positive-price and underflow safeguards below.\hfill$\square$

\paragraph{Proof of Proposition~\ref{prop:scalarization}.}
With zero uncertainty, $Z_k=0$. For $h=0$, the excess terminal value is $V_0(x)=x_+$. If $V_{h-1}(x)=\max\{0,x-(h-1)\gamma\}$, then
\begin{equation}
 V_h(x)=\max\{0,-\gamma+\max\{0,x-(h-1)\gamma\}\}
       =\max\{0,x-h\gamma\}.
\end{equation}
Thus the continuation root is $h\gamma$. With quality terminal utility $Q_i$, the index is $G_i^Q=Q_i-h\eta_Q\kappa_i=Q_i-\eta_QhBD_i$. With cost terminal utility $C_i=1-D_i/R$, the same recursion gives $G_i^C=C_i-h\eta_C\kappa_i=1-(R^{-1}+\eta_ChB)D_i$. Since $R>0$ and $\eta_C>0$, the cost-axis coefficient is positive. Required completion is essential: if a known terminal reward could be collected immediately, paying the remaining $h$ batches would be unnecessary. Unequal remaining horizons instead give configuration-specific penalties $\eta_a h_iBD_i$, not a common scalarization weight.\hfill$\square$

\section{Posterior Calibration and Search Implementation}
\label{app:posterior}
\label{app:dp}

Warm-start batches use independent seeded question orders for each configuration. The prior means are equal-weight averages of configuration warm-batch means. Quality prior variance is the sample variance of configuration warm-batch quality means, floored at $10^{-12}$. If $\bar D_i^{\mathrm{warm}}$ is the warm-batch mean cost and $K>1$, use
\begin{equation}
 v_{0,D}=\max\left\{\frac1{K-1}\sum_{i=1}^K
  \left(\bar D_i^{\mathrm{warm}}-\frac1K\sum_{j=1}^K\bar D_j^{\mathrm{warm}}\right)^2,
  10^{-12}R^2\right\}.
\end{equation}
Freeze calibration after warm-up and include each warm observation in the likelihood once. Empirical-Bayes reuse of warm data and the independent Gaussian approximation are modeling choices, not coverage certificates.

Equation~\eqref{eq:scalar-dp} is one-dimensional on either exact axis. Only the active objective's latent-posterior-mean innovation variance enters its scalar dynamic program; the unused objective is still observed and updates the shared posterior. Root tables are cached by variance schedule, horizon, and positive effective cost. The quality and cost axes therefore do not require the expected-minimum formula or a two-dimensional radial dynamic program. The cached tables plan full-batch posterior updates; a shorter final batch uses its actual size for the observation update and API accounting. Proposition~\ref{prop:scalarization} assumes equal-sized, equal-price remaining batches. For acquisition, the reported implementation rounds $B$ times each configuration's warm-start mean deployment cost to the nearest multiplicative bin with ratio $2$ and anchor $10^{-4}$ USD, then applies $\eta_a$; these frozen prices allow table reuse. The deployment-cost posterior and realized-dollar accounting continue to update.

\paragraph{Controller sets and Pareto recommendation.}
Algorithm~\ref{alg:cc} covers both objective settings. For quality--cost ($d=2$), the controller set is $\mathcal A=\{C,Q\}$, visited in the order $C,Q$, with terminal utilities $C_i$ and $Q_i$. For quality--cost--latency ($d=3$), use $\mathcal A=\{Q,\mathrm{bal},\ell,C\}$, visited in the order $Q,\mathrm{bal},\ell,C$. Here $\ell_i=1-L_i/T$, with $T>0$ frozen after warm-up, and the balanced terminal utility is $\min\{Q_i,\ell_i\}$. Pure controllers use one-dimensional indices; the balanced controller uses the two-coordinate radial dynamic program described in Appendix~\ref{app:extension}. Every controller has its own positive multiplier $\eta_a$ and shares the observed outcomes.

The conservative Pareto recommendation rule returns configurations that are nondominated when maximizing $Q^-$ and minimizing $D^+$ from Eq.~\eqref{eq:pareto-recommendation}. For $d=3$, it also minimizes $L^+_{i,n}=M_{i,L,n}+\beta\sqrt{U_{i,L,n}}$, with the same $\beta=1$. Thus each purchased batch updates two or three posterior marginals, and the recommendation is formed in the corresponding objective space.

\paragraph{Numerical stopping rule.}
The replay does not impose a fixed lower bound on $\eta_a$. Let $G^a_{\max}$ be the best unfinished index; the implementation treats $G^a_{\max}\le I_a+\tau$ as a local stopping signal. At such a stop, let $\rho_a=\max_{i\text{ unfinished}}h_i\gamma_{i,a}$ be the largest remaining effective completion penalty, where $h_i$ is the number of planned batches remaining and $\gamma_{i,a}$ is the controller's current effective per-batch price. The numerical threshold is
\begin{equation}
 \delta_a=\max\{\tau,\epsilon_{\mathrm{mach}}\max(1,|I_a|,|G^a_{\max}|)\},
 \label{eq:eta-numerical-threshold}
\end{equation}
where $\epsilon_{\mathrm{mach}}$ is double-precision machine epsilon and the available implementation defaults to $\tau=10^{-9}$. If $\rho_a>\delta_a$ and halving keeps the multiplier and effective costs positive, the visited controller halves $\eta_a$ and yields without an evaluation. Otherwise, it records a numerical stop, including when the next halving would underflow. Let $F$ collect the current numerical-stop marks. A newly purchased batch clears $F$. Acquisition ends when every controller in $\mathcal A$ is marked without an intervening observation, an optional external cost cap is reached, or all configurations are complete. Users may also stop after any completed batch and retain the latest recommendation. A numerical stop is not statistical certification. Algorithm~\ref{alg:cc} gives the replay rule.

\begin{algorithm}[H]
\caption{CC-Gittins replay for quality--cost and quality--cost--latency}
\label{alg:cc}
\begin{algorithmic}[1]
\Require configurations, objective setting $d\in\{2,3\}$, controller set $\mathcal A$ and visit order, batch size $B$, optional external API cost cap, tolerance $\tau$
\State acquire warm batches; freeze posterior calibration, objective scales, and $\widehat\kappa_i$
\State set $\eta_a=1$ for every $a\in\mathcal A$ and $F=\varnothing$
\State publish the conservative Pareto set in the selected $d$ objectives
\While{search has not been interrupted, any external cost cap permits evaluation, unfinished configurations remain, and $F\ne\mathcal A$}
  \State $a\gets$ next controller in the specified round-robin order
  \State $i\gets\arg\max_{j\text{ unfinished}}G^a_{j,k_j}$; obtain completed incumbent $I_a$
  \If{$G^a_{i,k_i}\le I_a+\tau$}
    \State compute remaining penalty $\rho_a$ and threshold $\delta_a$
    \If{$\rho_a>\delta_a$ and the next halving keeps effective costs positive}
      \State $\eta_a\gets\eta_a/2$; remove $a$ from $F$
    \Else
      \State add $a$ to $F$
    \EndIf
    \State \textbf{continue} to the next visit
  \EndIf
  \State acquire $i$'s next batch, charge realized API cost, and update all $d$ posterior marginals
  \State $F\gets\varnothing$; publish the updated conservative Pareto set
\EndWhile
\State \Return the latest published Pareto recommendation
\end{algorithmic}
\end{algorithm}

\section{Additional Objectives and the Quality--Cost--Latency Extension}
\label{app:extension}

Suppose $d\ge2$ objectives include per-query deployment cost $D_i$, and let $z_i\in\R^{d-1}$ collect the remaining higher-is-better objectives. For $v\in\Delta^{d-2}$, use terminal utility $U_v(i)=v^\top z_i$. Under zero uncertainty, equal remaining horizon $h$ of full batches, and exact coupling $\kappa_i=BD_i$, the index ranks
\begin{equation}
 v^\top z_i-\eta hBD_i=v^\top z_i-\lambda D_i.
\end{equation}
A normalized linear weight vector over $d$ objectives has $d-1$ degrees of freedom. Cost coupling supplies the cost weight, leaving $d-2$ explicit direction dimensions among non-cost objectives. This statement concerns linear scalarizations with positive total weight on non-cost objectives; a pure cost controller handles the remaining cost extreme. It does not remove unsupported-front geometry or guarantee that a discrete $\eta$ schedule realizes every weight.

\paragraph{Quality--cost--latency controllers.}
Let $L_i$ be the mean per-question sum of server-reported LLM request latencies, and set $\ell_i=1-L_i/T$ using a positive scale $T$ frozen after warm-up. The reported method has four directions in acquisition coordinate order $(Q,C,\ell)$: pure quality $(1,0,0)$, balanced quality--latency $(\tfrac12,0,\tfrac12)$, pure latency $(0,0,1)$, and pure cost $(0,1,0)$. The pure directions use the corresponding scalar desirabilities. The balanced direction instead uses the \emph{radial} terminal utility $\min\{Q_i,\ell_i\}$ in Eq.~\eqref{eq:three-objective-index}; its direction specifies equal scaling of two desirabilities, not their arithmetic mean. The choice of this intermediate direction is empirical. In particular, the linear supported-front argument above does not prove that these four controllers cover the three-objective frontier.

\paragraph{Posterior and acquisition.}
The three raw outcomes have independent Gaussian posterior marginals. Quality and cost use the scalar indices in Eq.~\eqref{eq:scalar-dp}; latency uses the latent-posterior transform $m^\ell_{i,n}=1-m_{i,L,n}/T$ and $v^\ell_{i,n}=v_{i,L,n}/T^2$. For the balanced direction, the two-coordinate radial dynamic program uses the latent posterior means and variances of $(Q_i,\ell_i)$ and the expected minimum as its terminal reward. The finite-benchmark-mean bridge in Eq.~\eqref{eq:finite-bridge} is used for recommendation. The balanced controller does not project a joint posterior onto the linear average $(Q_i+\ell_i)/2$, and it makes no fitted cross-objective covariance claim. Each measured execution updates all three outcomes for every controller. Controllers take turns, independently relax their positive evaluation-cost penalties after local stopping signals, and charge observed token-priced API spend; latency does not price evaluation.

\paragraph{Recommendation and evaluation geometry.}
Augment Eq.~\eqref{eq:pareto-recommendation} with $L^+_{i,n}=M_{i,L,n}+\beta\sqrt{U_{i,L,n}}$ and return configurations nondominated when maximizing $Q^-$ and minimizing $D^+,L^+$. This conservative recommendation rule retains the same heuristic status as its two-objective counterpart; the random baselines form empirical three-objective Pareto recommendations. For offline evaluation only, the true means are mapped to
\begin{equation}
 y_i^{(3)}=\left(Q_i,\frac{R_L}{R_L+L_i},\frac{R_D}{R_D+D_i}\right),
 \label{eq:three-objective-normalization}
\end{equation}
where $R_L$ and $R_D$ are medians of strictly positive full-data assignment means for latency and deployment cost, respectively. These nonlinear reciprocal scales never enter acquisition. Using the origin reference, we compute three-dimensional HV regret and Euclidean GD/IGD on these fixed coordinates. The complete three-dimensional frontiers have 11 points on MathQA and 29 on HotpotQA; we did not compute a three-dimensional supported-front oracle.

\paragraph{Experimental integration.}
The two three-objective replays use the complete HotpotQA and MathQA matrices (100 assignments and 200 questions each), four warm questions per assignment, batches of up to four unseen questions, and 20 paired question-order seeds. We use cumulative server-reported LLM latency to reduce sensitivity to client-side processing and network conditions. For each configuration, we average the summed Bedrock-reported durations of successful LLM calls over all questions. Tool execution time is outside the scope of this metric. Optimizer overhead is measured separately. On a fixed 0.5-percentage-point grid, we show each seed's latest completed checkpoint at or below the target fraction of exhaustive observed API spend, without interpolating or overspending. At each target, means and $\pm 2$ sample-standard-error bands summarize the contributing seeds, and band lower bounds are clipped at zero. The random-question baseline evaluates a shared question prefix across assignments, while random configurations evaluates complete assignment prefixes; both form empirical three-objective Pareto recommendations. These comparisons therefore include differences in both evaluation selection and recommendation rules. Figure~\ref{fig:three-objective-four} also includes three-objective EGE-SH, APE-$k$, and qNEHVI.

\paragraph{Controlled ablations.}
Appendix~\ref{app:three-objective-ablations} compares real and unit acquisition prices, single-axis and multi-direction controller sets, direction-local relaxation against fixed-price continuation, and prior-variance calibration. These controls retain the same three-objective recommendation rule and realized-dollar accounting. Their results distinguish improvements in dominated-volume recovery and recommendation proximity from improvements in frontier coverage.

\section{Experimental Setup and Implementation Details}
\label{app:implementation}

\subsection{Benchmark Matrices}
\label{app:benchmark-matrices}

Each response matrix has one row per retained role-level model assignment and one column per question. We collect complete HotpotQA and MathQA matrices by evaluating all $10^2=100$ assignments of ten candidate models to two roles on all 200 questions, or 20,000 observed cells per benchmark. The two workflow settings follow those described in AgentOpt~\citep{hua2026agentopt}. The calculated exhaustive API costs are \$62.58 and \$234.34, respectively. Their full-data Pareto frontiers contain 15 and 6 assignments, and their maximum mean quality scores are 74.66\% token-level F1 for HotpotQA and 97.50\% accuracy for MathQA. The six SCOPE matrices use fully evaluated configuration rows from SCOPE's released per-query cache for three data-processing systems~\citep{huang2026scope}. Every retained SCOPE configuration has score and token records for all questions in its corresponding benchmark split. Summing the per-entry API costs over each retained configuration--question matrix gives calculated exhaustive API costs of \$2,067.38, \$946.78, \$149.24, \$14.84, \$80.15, and \$2.55 for BIRD Mini Dev, BIRD Dev, Bing Query Logs, Stack Overflow, Restaurant Valid, and Restaurant Test, respectively. These configurations cover a subset of the nominal Cartesian product of role-level model choices; we do not infer results for unobserved configurations. Table~\ref{tab:benchmark-inventory} reports the retained dimensions used by every compared method. Thus, the exhaustive-evaluation cost and full-data Pareto frontier for each instance refer to its retained matrix; paired development and test instances may contain different assignment sets.

\begin{table}[H]
  \centering
  \caption{Benchmark matrices used in the eight-instance evaluation. We collected HotpotQA and MathQA observations with our workflow evaluation harness; the other six instances use SCOPE's released evaluation records. ``Configurations'' counts retained role-level model assignments (the arms of the search problem), and ``Questions'' counts each instance's evaluation horizon, covered by every retained configuration.}
  \label{tab:benchmark-inventory}
  \small
  \setlength{\tabcolsep}{5pt}
  \begin{tabular}{@{}llrrr@{}}
    \toprule
    Benchmark instance & Workflow & Roles & Configurations & Questions \\
    \midrule
    HotpotQA & Planner--solver & 2 & 100 & 200 \\
    MathQA & Answerer--critic & 2 & 100 & 200 \\
    BIRD Mini Dev & DIN-SQL & 4 & $1{,}115$ & 500 \\
    BIRD Dev & DIN-SQL & 4 & 75 & $1{,}534$ \\
    Bing Query Logs & UniDM-DT & 5 & $1{,}681$ & 102 \\
    Stack Overflow & UniDM-DT & 5 & 23 & 710 \\
    Restaurant Valid & UniDM-DI & 3 & 972 & 156 \\
    Restaurant Test & UniDM-DI & 3 & 36 & 86 \\
    \bottomrule
  \end{tabular}
\end{table}

At the question level, HotpotQA uses a continuous token-level F1 score in $[0,1]$. Since HotpotQA produces free-form answers, token-level F1 preserves partial credit for token overlap that binary exact match would discard. MathQA uses a binary correctness indicator, and each released per-query evaluation in the six SCOPE-derived instances likewise provides a binary task-success outcome. Consequently, configuration-level mean quality is mean F1 for HotpotQA and the empirical success rate for the other seven instances.

\subsection{Model Pricing and Cost Accounting}
\label{app:agentopt-pricing}

The ten candidate models use AWS Bedrock Application Inference Profiles. Table~\ref{tab:agentopt-model-prices} lists the ten models in the HotpotQA and MathQA matrices and the AWS Bedrock US standard on-demand prices used for cost accounting.\footnote{\url{https://aws.amazon.com/bedrock/pricing/}; \url{https://docs.aws.amazon.com/bedrock/latest/userguide/model-card-openai-gpt-56-luna.html}.} Each workflow execution's API cost is the sum, across both roles, of its recorded input and output token counts multiplied by the corresponding per-million-token rates. The GPT-5.6 Luna rate uses the US commercial short-context tier (up to 272K input tokens per request). The six SCOPE-derived matrices use the token records and model API prices in their released evaluation cache.

\begin{table}[H]
  \centering
  \caption{AWS Bedrock US standard on-demand input and output prices used for the ten models in our HotpotQA and MathQA evaluations, in USD per million tokens.}
  \label{tab:agentopt-model-prices}
  \small
  \begin{tabular}{@{}lrr@{}}
    \toprule
    Model & Input & Output \\
    \midrule
    Claude Haiku 4.5 & 1.00 & 5.00 \\
    Claude Opus 4.6 & 5.00 & 25.00 \\
    DeepSeek R1 & 1.35 & 5.40 \\
    GPT-5.6 Luna & 0.22 & 1.32 \\
    gpt-oss-120b & 0.15 & 0.60 \\
    gpt-oss-20b & 0.07 & 0.30 \\
    Kimi K2.5 & 0.60 & 3.00 \\
    Ministral 3 8B & 0.15 & 0.15 \\
    Qwen3 32B & 0.15 & 0.60 \\
    Qwen3 Next 80B A3B & 0.15 & 1.20 \\
    \bottomrule
  \end{tabular}
\end{table}

\subsection{Two-objective CC-Gittins configuration}

In practice, evaluation queries can be run in batches of size $B$ to reduce wall-clock latency through parallel execution; $B=1$ is also valid.

\begin{table}[H]
\centering\small
\caption{Two-objective CC-Gittins settings for the reported replay.}
\label{tab:implementation-settings}
\begin{tabular}{@{}ll@{}}
\toprule
Component & Setting \\
\midrule
Warm start / adaptive batch & 4 / up to 4 questions per configuration \\
Statistical target & Mean of the fixed benchmark \\
Cost coordinates & Deployment cost (USD per query); $C=1-D/R$ for acquisition \\
Directions (in visit order) & $(0,1)$ cost, $(1,0)$ quality \\
Scheduler & 1:1 decision opportunities, shared observations \\
Multipliers & $\eta_C=\eta_Q=1$; independent halving \\
Numerical safeguard & Precision threshold in Eq.~\eqref{eq:eta-numerical-threshold}; underflow check \\
Termination & Both local numerical stops, replay cost cap, or exhaustive completion \\
Recommendation & Conservative Pareto, $\beta=1$, no set-size cap \\
Search cost accounting & Token-based API cost (USD), including warm start \\
\bottomrule
\end{tabular}
\end{table}

\subsection{Baseline implementations}
\label{app:baselines}

All five baselines use the same frozen response matrices but follow their own sampling protocols; checkpoints are aligned by realized cumulative API search cost. EGE-SH allocates observations in rounds and uses empirical Pareto gaps to halve the active assignments~\citep{kone2024bandit}. We use the confidence-bound sampling rule of APE-$k$ to direct observations toward uncertain Pareto comparisons; its fixed-confidence stopping condition is disabled, and sampling continues until the replay observation cap is reached or available observations are exhausted~\citep{kone2023adaptive}. qNEHVI uses a Bayesian surrogate over categorical assignment features and chooses evaluations by noisy expected hypervolume improvement~\citep{daulton2021parallel}.

The two random search baselines differ in the sampled unit: \emph{random questions} evaluates all assignments on the same randomly drawn questions, while \emph{random configurations} fully evaluates randomly drawn assignments. Both recommend the empirical Pareto set from their observed outcomes.

\clearpage
\section{Supported-Front Analysis}
\label{app:supported-front-diagnostic}

The full-data supported-front analysis examines what the weighted-score family can represent on each matrix. For this analysis, HV and distances use $(Q_i,1-D_i/(1.05\max_j D_j))$, with the origin as the HV reference. For HotpotQA, 6 of the 15 true Pareto points are supported. Keeping those six loses only $0.00066$ of absolute hypervolume, although the furthest omitted frontier point is $0.0716$ away in normalized coordinates. Stack Overflow and Restaurant Test have no unsupported frontier points. This offline oracle considers all nonnegative weights and known outcomes; it does not assess \ccgittins{} recommendations at intermediate search costs.

\section{CC-Gittins Replay Results}
\label{app:cc-gittins-results}

\paragraph{Protocol and scope.}
\ccgittins{} uses the exact cost and quality axes $(0,1)$ and $(1,0)$, real per-configuration acquisition costs, independent direction-local $\eta$ decay, shared observations, and round-robin controller visits. The evaluation covers eight benchmarks and seeds 42--61. HotpotQA and MathQA use the complete $10\times10$ assignment matrices.

The $0$--$30\%$ curves use fixed 0.5-percentage-point targets and summarize the latest recommendation recorded at or below each target.

\subsection{Recommendation snapshots at cost checkpoints}

Figures~\ref{fig:pareto-10} and~\ref{fig:pareto-30} in Appendix~\ref{app:additional-plots} show recommendation snapshots at cost checkpoints of 10\% and 30\% of exhaustive API evaluation cost across eight instances. Each snapshot is the latest recommendation available at or below the checkpoint on the recorded search trajectory. At 10\%, the recommendation contains a small set of configurations supported by the available observations. By 30\%, more configurations have narrower predictive intervals, and the set traces more of the quality--cost frontier. The quality controller can favor interior supported trade-offs through its cost penalty, while the cost controller directly explores the low-cost region. Figures~\ref{fig:estimated-actual-seed42-10} and~\ref{fig:estimated-actual-seed42-30} then illustrate the recommended configurations and quality--cost estimates available to users under \ccgittins{} and the five baselines for seed 42.

\subsection{Ablations and interpretation}

Appendix~\ref{app:ablations} varies one conceptual component around G0: acquisition-cost pricing, controller directions, relaxation, scheduling, or prior calibration. At the 30\% checkpoint, G0 has lower mean HV regret, GD, and IGD than every cost-mechanism control. Adding a midpoint or five directions can lower HV regret or GD further, but neither improves all three metrics: G0 retains the lowest mean IGD in that family. The fixed-$\eta$ no-stop control has worse HV regret, GD, and IGD than G0's relaxation mechanism, while sequentially draining one axis underperforms round-robin independent relaxation. We report HV, GD, and IGD together since improvements in dominated area, recommendation reliability, and frontier coverage need not coincide.

\section{Runtime Results}\label{app:runtime}

Figure~\ref{fig:wall-clock-runtime} compares the total wall-clock runtime of the six methods. qNEHVI has the highest mean runtime on all eight benchmarks, and the gap widens substantially on the three largest SCOPE matrices. Its mean runtime reaches 6.2 hours on Restaurant Valid, 10.4 hours on BIRD Mini Dev, and 22.5 hours on Bing Query Logs, compared with 1.1, 3.6, and 3.0 minutes for \ccgittins{}, respectively. \ccgittins{} has a mean runtime below 3.6 minutes on every benchmark. EGE-SH and APE-$k$ also remain below 18 minutes, while the two random baselines finish in approximately 8--14 seconds. These results show that qNEHVI's computational overhead becomes the dominant cost as the configuration matrix grows.

\begin{figure}[H]
  \centering
  \includegraphics[width=0.98\linewidth]{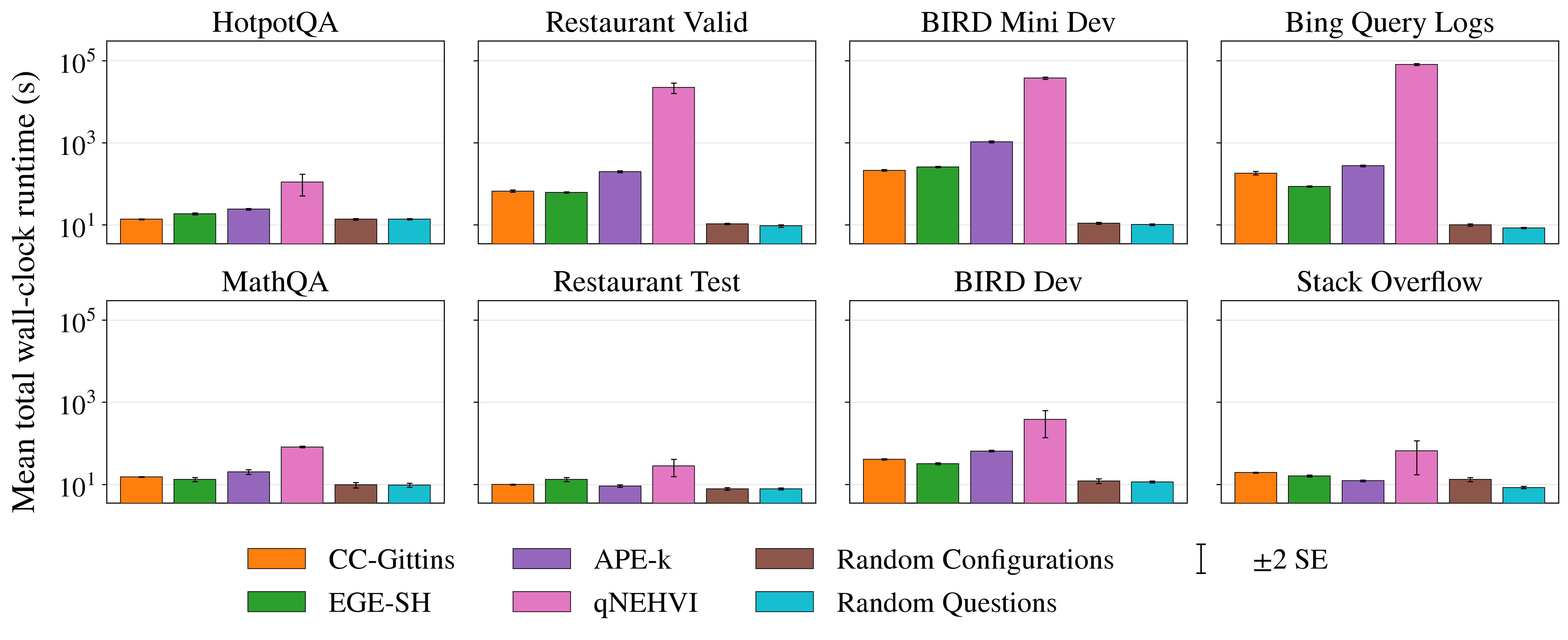}
  \caption{Mean total wall-clock runtime of the six methods across eight benchmark instances. The vertical axis uses a base-10 logarithmic scale, with major ticks marking successive powers of ten. Bars show means and error bars show $\pm2$ standard errors across 20 matched seeds. CC-Gittins timings cover the complete child Python process, including startup, data loading, policy computation, postprocessing, and result serialization; parent validation and Slurm queue time are excluded.}
  \label{fig:wall-clock-runtime}
\end{figure}

\section{CC-Gittins Pareto Plots}
\label{app:additional-plots}

\begin{figure}[H]
  \centering
  \includegraphics[width=0.98\linewidth]{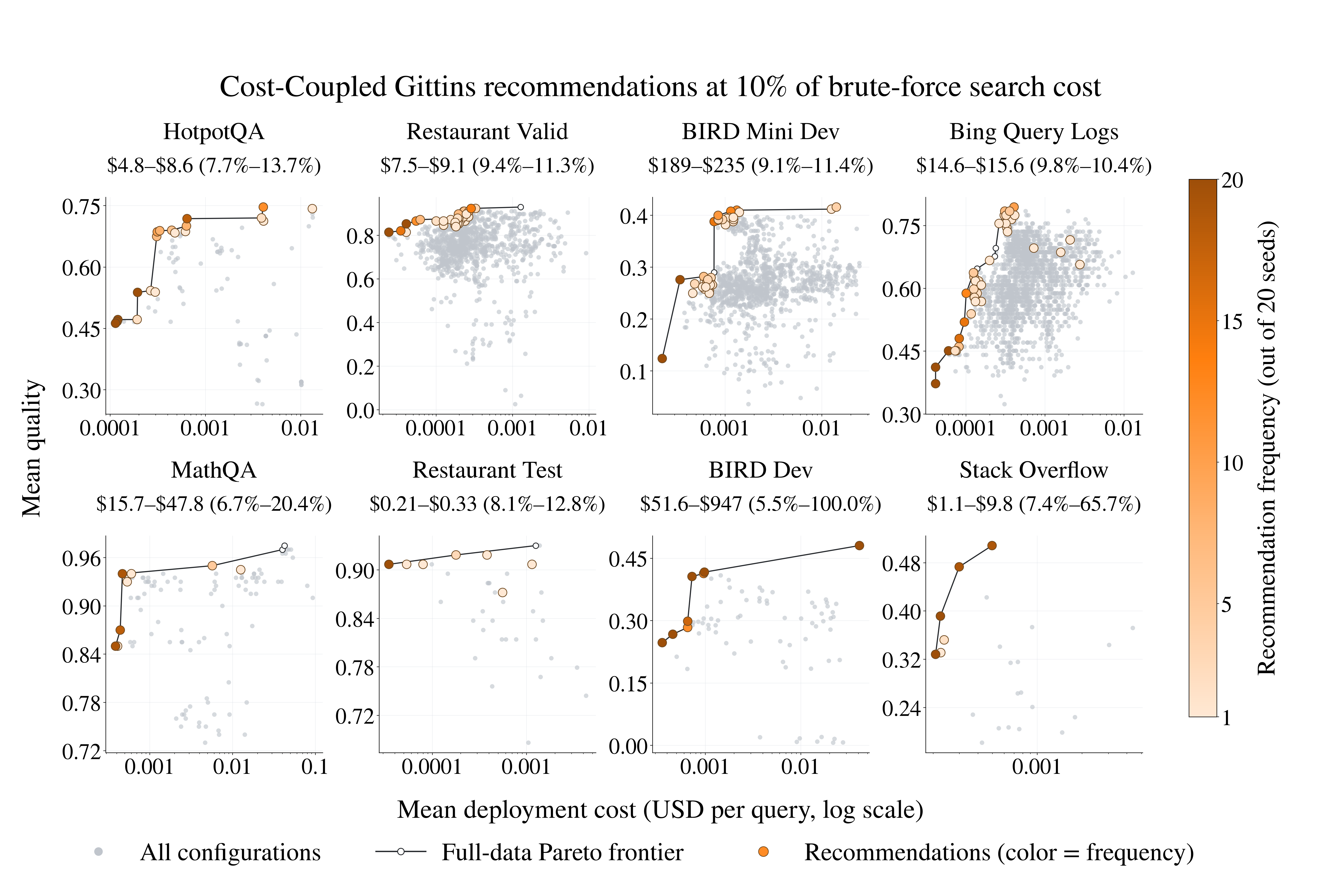}
  \caption{\ccgittins{} recommendation snapshots at the 10\% checkpoint of exhaustive API evaluation cost across eight benchmark instances and 20 seeds. Each snapshot uses the latest recommendation available at or below the checkpoint. Gray points show all configurations; the black line shows the full-data Pareto frontier. Orange markers locate recommended configurations at their exhaustive mean quality and deployment cost, with color indicating recommendation frequency across seeds. The line beneath each panel title reports the mean start--end interval over which the checkpoint recommendation remains stable, first as cumulative API cost (USD) and then as a fraction of exhaustive search cost.}
  \label{fig:pareto-10}
\end{figure}

\begin{figure}[H]
  \centering
  \includegraphics[width=0.98\linewidth]{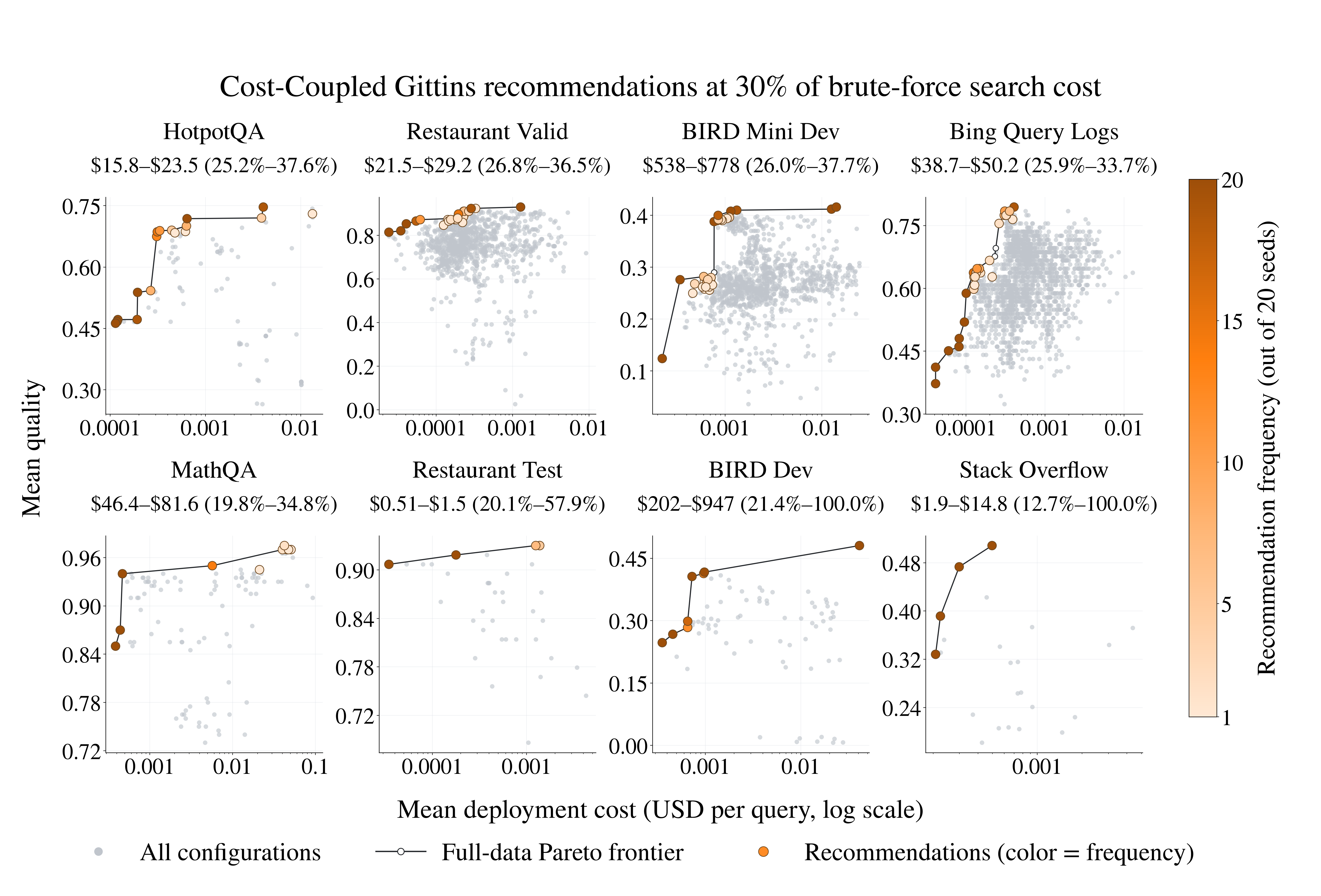}
  \caption{\ccgittins{} recommendation snapshots at the 30\% checkpoint of exhaustive API evaluation cost, displayed as in Fig.~\ref{fig:pareto-10}. Marker color gives recommendation frequency across 20 seeds; positions use exhaustive mean quality and deployment cost for retrospective comparison.}
  \label{fig:pareto-30}
\end{figure}

\section{Pareto Plot Protocol at Cost Checkpoints}
\label{app:plot-protocol}

For every benchmark, seed, and target $b\in\{0.1,0.3\}$, we select the last recommendation event whose realized cost fraction does not exceed $b$. The selected configurations are then placed at exhaustive raw mean quality and mean deployment cost (USD per query). These exhaustive coordinates are used only for offline evaluation and are not available to the selector.

The plot distinguishes completed from partial recommendations since their online semantics differ. A completed point has zero predictive variance for the benchmark mean and its displayed coordinate matches its recommendation estimate. A partial point is selected using Eq.~\eqref{eq:pareto-recommendation}, while its displayed exhaustive coordinate reveals estimation error. All methods use the same axis limits within a benchmark.

Figures~\ref{fig:estimated-actual-seed42-10} and~\ref{fig:estimated-actual-seed42-30} illustrate the recommended configurations and quality--cost estimates available to users, comparing \ccgittins{} with five baselines on HotpotQA and Stack Overflow at seed 42. Within each benchmark, the actual and estimated columns contain the same recommended configuration identifiers; only their plotted coordinates change. For \ccgittins{}, estimated coordinates are the full-benchmark posterior predictive means from Eq.~\eqref{eq:finite-bridge}; the baselines use means over their evaluated questions. The actual panels and exhaustive reference points and frontier serve only as retrospective references.

\begin{figure}[p]
  \centering
  \includegraphics[width=\textwidth]{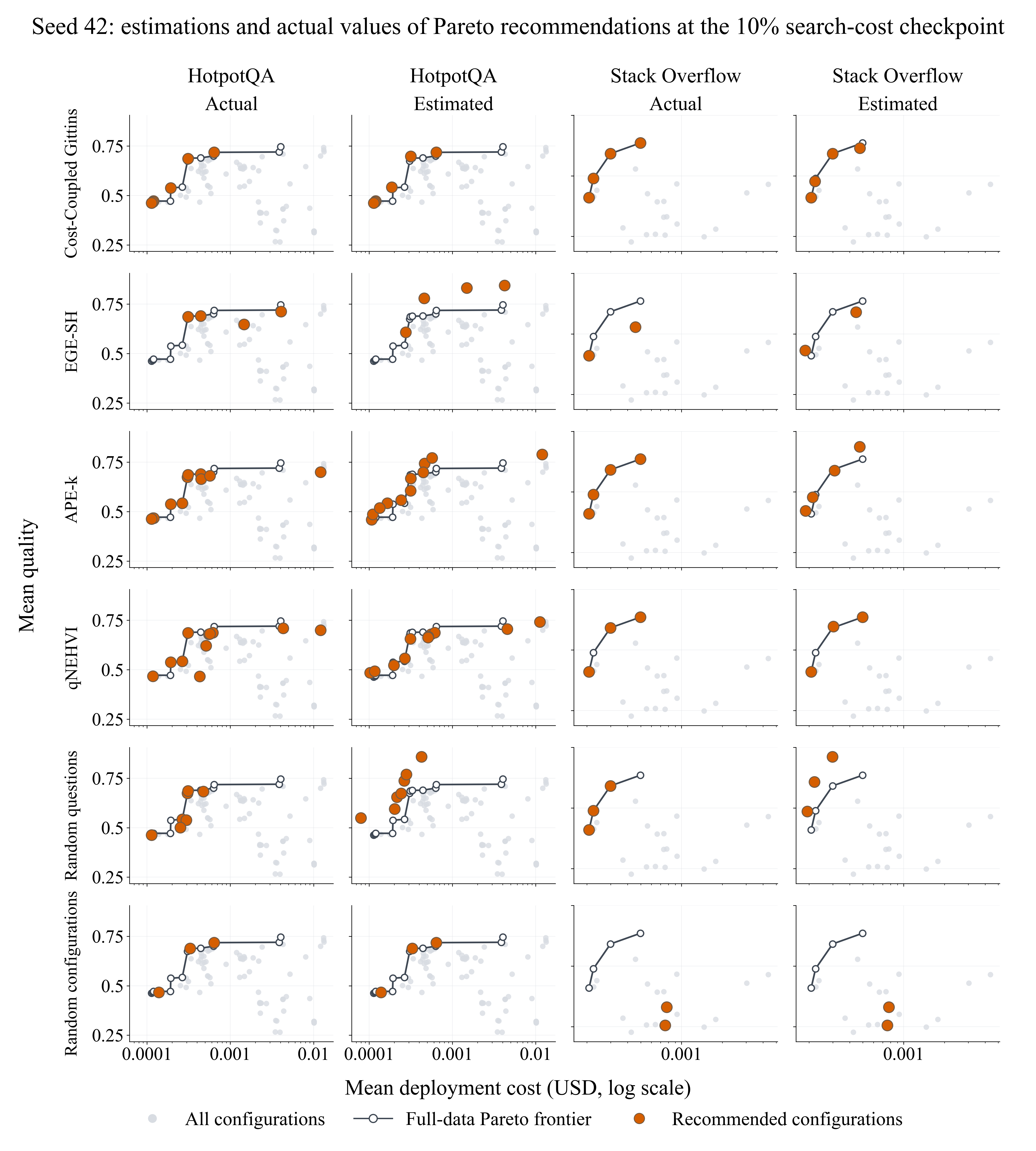}
  \caption{Recommendations and quality--cost estimates available to users at the 10\% search-cost checkpoint for seed 42. Rows show \ccgittins{} and the five baselines; columns show exhaustive (actual) and checkpoint-estimated coordinates for the same recommended configurations on HotpotQA and Stack Overflow. The \ccgittins{} estimates combine observed outcomes with posterior predictions for unobserved questions; baseline estimates are observed-question means. Gray points are all configurations, hollow markers and connecting lines show the exhaustive Pareto frontier, and orange points show the configurations recommended by each method.}
  \label{fig:estimated-actual-seed42-10}
\end{figure}

\begin{figure}[p]
  \centering
  \includegraphics[width=\textwidth]{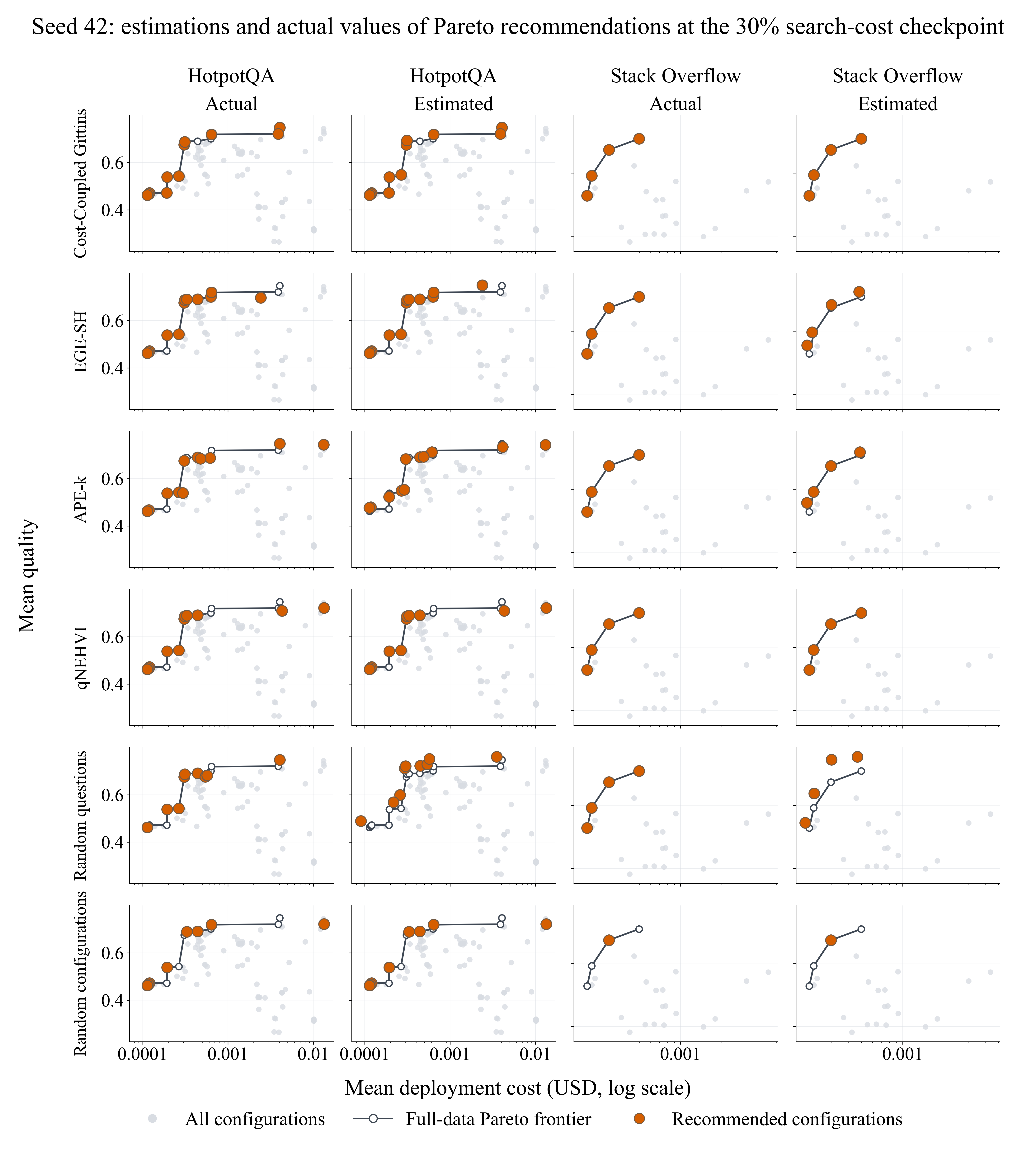}
  \caption{Recommendations and quality--cost estimates available to users at the 30\% search-cost checkpoint for seed 42, using the same layout and visual encoding as Fig.~\ref{fig:estimated-actual-seed42-10}. Rows compare \ccgittins{} with the five baselines; actual coordinates provide a retrospective reference for the same recommended configurations.}
  \label{fig:estimated-actual-seed42-30}
\end{figure}

\clearpage
\section{Controlled CC-Gittins Ablations}
\label{app:ablations}

\paragraph{Protocol.}
We center four one-factor families and a $2\times2$ prior-variance study (Fig.~\ref{fig:ablation-prior-variance}) on G0, the \ccgittins{} protocol with real per-configuration acquisition costs, independent direction-local $\eta$ decay, and round-robin axis visits. The cost family compares G0 with quality-only real cost (G1), quality-only unit cost (G2), and exact axes with unit cost (G3). The direction family includes quality-only G1, deployment-only G4, G0, axes plus the midpoint (G5), and five evenly spaced directions (G6), all with real costs. The relaxation family compares G0's independent direction-local relaxation with the fixed-$\eta$ no-stop control (G7). The scheduler family compares G0 with quality-then-deployment (G8) and deployment-then-quality (G9), which drain one axis before visiting the other.

Each configuration uses the same eight benchmark matrices, 20 seeds, model ordering, exhaustive truth vectors, recommendation rule, batch size, and cost checkpoints spanning $0$--$30\%$ of exhaustive API evaluation cost. The frontier-metric grid contains thirteen configurations and 2,080 validated benchmark--seed runs. At each fixed 0.5-percentage-point target, we score each contributing seed's last recommendation not exceeding that cost. Curves show seed means and bands show $\pm2$ standard errors. HV regret, GD, and IGD are all lower-is-better and use the same frozen per-benchmark metric space as the primary comparison.

\paragraph{Cost mechanism.}
Figure~\ref{fig:ablation-cost} separates acquisition pricing from
controller choice. G0 versus G3 and G1 versus G2 compare real with unit
prices; the comparisons retain deployment-cost observations, the Pareto recommendation rule,
and realized-dollar accounting. G0 versus G1 and G3 versus G2 test adding the cost
controller at fixed pricing. Real prices align acquisition with monetary
spend; the cost controller targets inexpensive assignments. These controls
support combining both mechanisms, with benchmark-dependent gains. They
assess pricing and controller choices within the coupled setting, rather
than isolating the effect of the relation $\kappa_i=BD_i$ itself.

\paragraph{Directions.}
Figure~\ref{fig:ablation-directions} shows that extra directions redistribute evaluation effort rather than uniformly improving the recommendation. Intermediate controllers can concentrate evidence on a compact set of high-value trade-offs, improving dominated-volume recovery and recommendation proximity while weakening uniform frontier coverage. The one-axis controls show the benefit of retaining both endpoint-oriented controllers, while adding directions does not mechanically enlarge the final set because all controllers share observations and a common conservative recommendation rule forms the returned set. The frontier-metric curves compare variants at matched fractions of exhaustive API evaluation cost, whereas the runtime experiment fixes the number of evaluated matrix entries at $\lceil0.30KN\rceil$, where $K$ is the number of configurations and $N$ is the number of questions in the benchmark matrix. The latter measures policy computation at equal evaluation counts rather than equal search cost. Including the single-direction controls, the runtime experiment covers five settings (Fig.~\ref{fig:ablation-directions-runtime}). G0 has lower mean policy runtime than both denser direction sets on all eight benchmarks. Averaging per-benchmark runtime ratios, G5 and G6 require 1.39$\times$ and 1.74$\times$ G0's policy time, respectively. Together with the frontier-quality controls, these results support the two-axis configuration as a trade-off between frontier recovery and policy computation.

\paragraph{Relaxation.}
Figure~\ref{fig:ablation-continuation} compares G0's independent direction-local relaxation with the fixed-$\eta$ no-stop control G7. In G0, a local stopping signal triggers a reduction of the visited controller's information-price multiplier when possible, allowing configurations whose value of information was marginal at an earlier price scale to become eligible for further evaluation. G7 instead retains the initial price scale and continues without a Gittins stop. G0 achieves lower aggregate mean HV regret, GD, and IGD at the 30\% checkpoint. These results support the relaxation mechanism for improving volume recovery and frontier coverage while maintaining competitive recommendation proximity; the gains vary across benchmarks.

\paragraph{Scheduler.}
Figure~\ref{fig:ablation-scheduler} shows that controller order affects recommendation quality at matched cumulative search cost: a sequential policy allocates early evaluations to its first controller and postpones evidence about the opposite frontier extreme. Round-robin scheduling instead balances directional attention from the outset, while shared observations allow every purchased batch to inform both controllers. This combination exposes both endpoints early enough for their evidence to shape the interior recommendation, making interleaving more robust than draining either axis first.

\clearpage
\subsection{Cost-mechanism curves}

\begin{figure}[H]
  \centering
  \includegraphics[width=\textwidth,height=0.76\textheight,keepaspectratio]{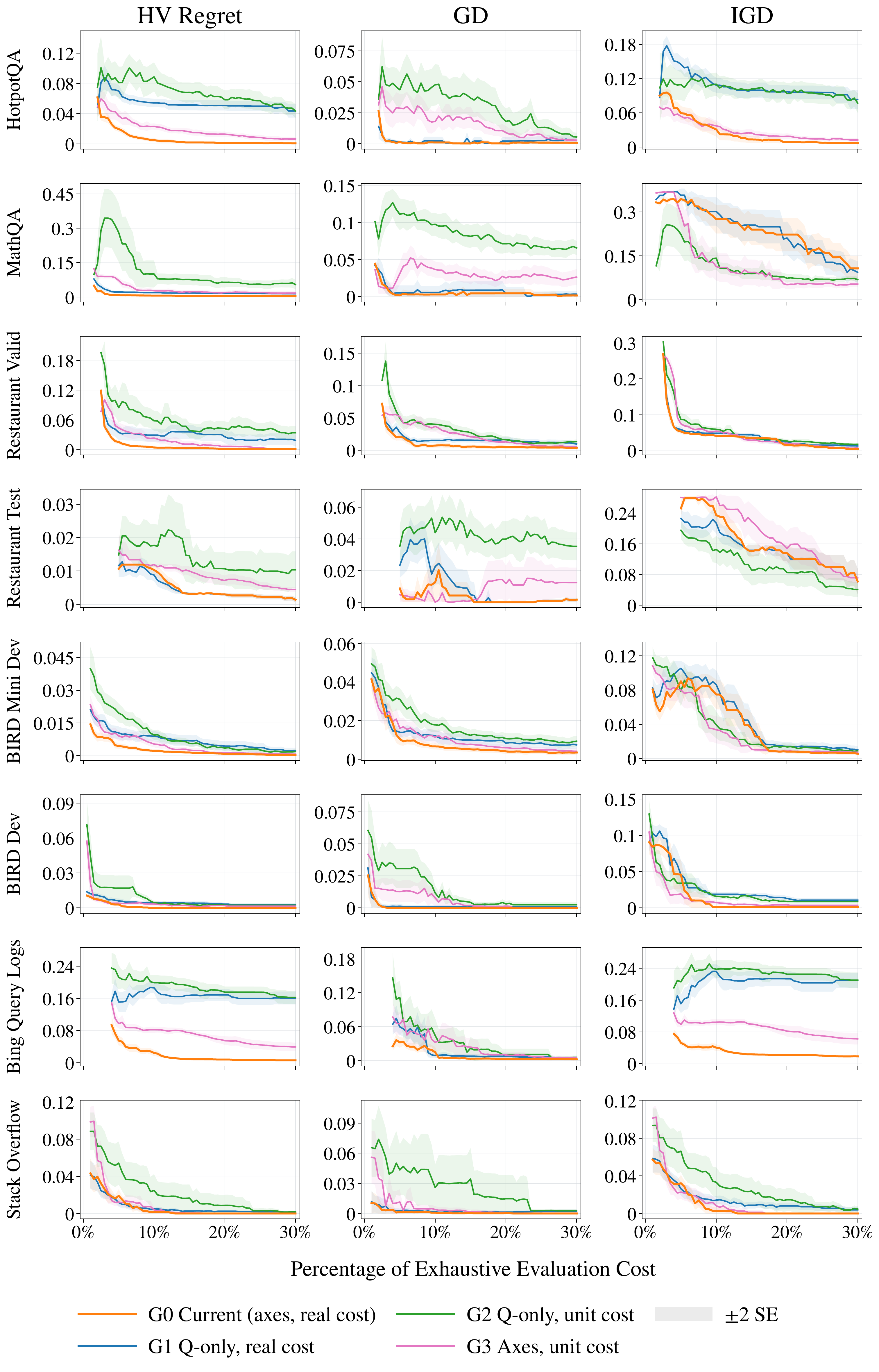}
  \caption{Cost-mechanism ablation across eight benchmarks (lower is better). Rows group the two-, three-, four-, and five-role workflows; columns report hypervolume (HV) regret, generational distance (GD), and inverted generational distance (IGD). G0 combines real acquisition costs with quality and deployment-cost controllers and achieves the lowest aggregate mean of all three metrics at the 30\% checkpoint. Its advantage over quality-only acquisition (G1, G2) and unit-cost axes (G3) is especially clear on several benchmarks, while the coverage advantage remains benchmark-dependent. Curves show seed means at fixed API cost checkpoints; shaded bands indicate $\pm2$ standard errors.}
  \label{fig:ablation-cost}
\end{figure}

\clearpage
\subsection{Direction curves}

\begin{figure}[H]
  \centering
  \includegraphics[width=\textwidth,height=0.76\textheight,keepaspectratio]{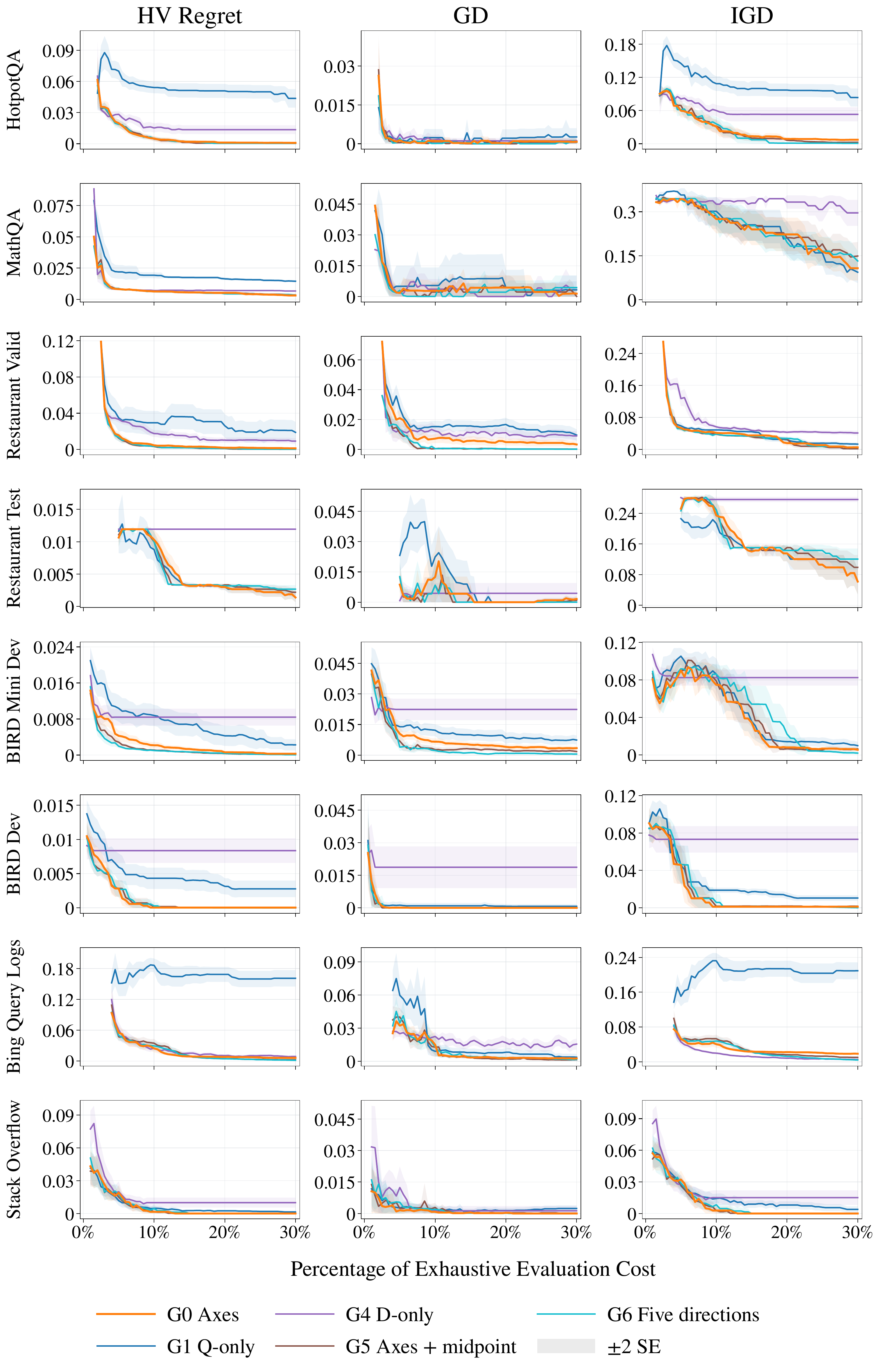}
  \caption{Direction ablation across eight benchmarks (lower is better). Columns report HV regret, GD, and IGD, and rows are ordered by workflow role count. G0 achieves competitive frontier metrics using only two exact-axis controllers and substantially improves on the single-direction controls on most benchmarks. At 30\%, G0 has the lowest aggregate mean IGD; adding directions does not uniformly improve all three frontier metrics. Together with G0's runtime relative to denser direction sets (Fig.~\ref{fig:ablation-directions-runtime}), these results support the two-axis design as an economical overall choice. All variants use real acquisition costs; curves show seed means and shaded bands indicate $\pm2$ standard errors.}
  \label{fig:ablation-directions}
\end{figure}

\begin{figure}[p]
  \centering
  \includegraphics[width=\textwidth]{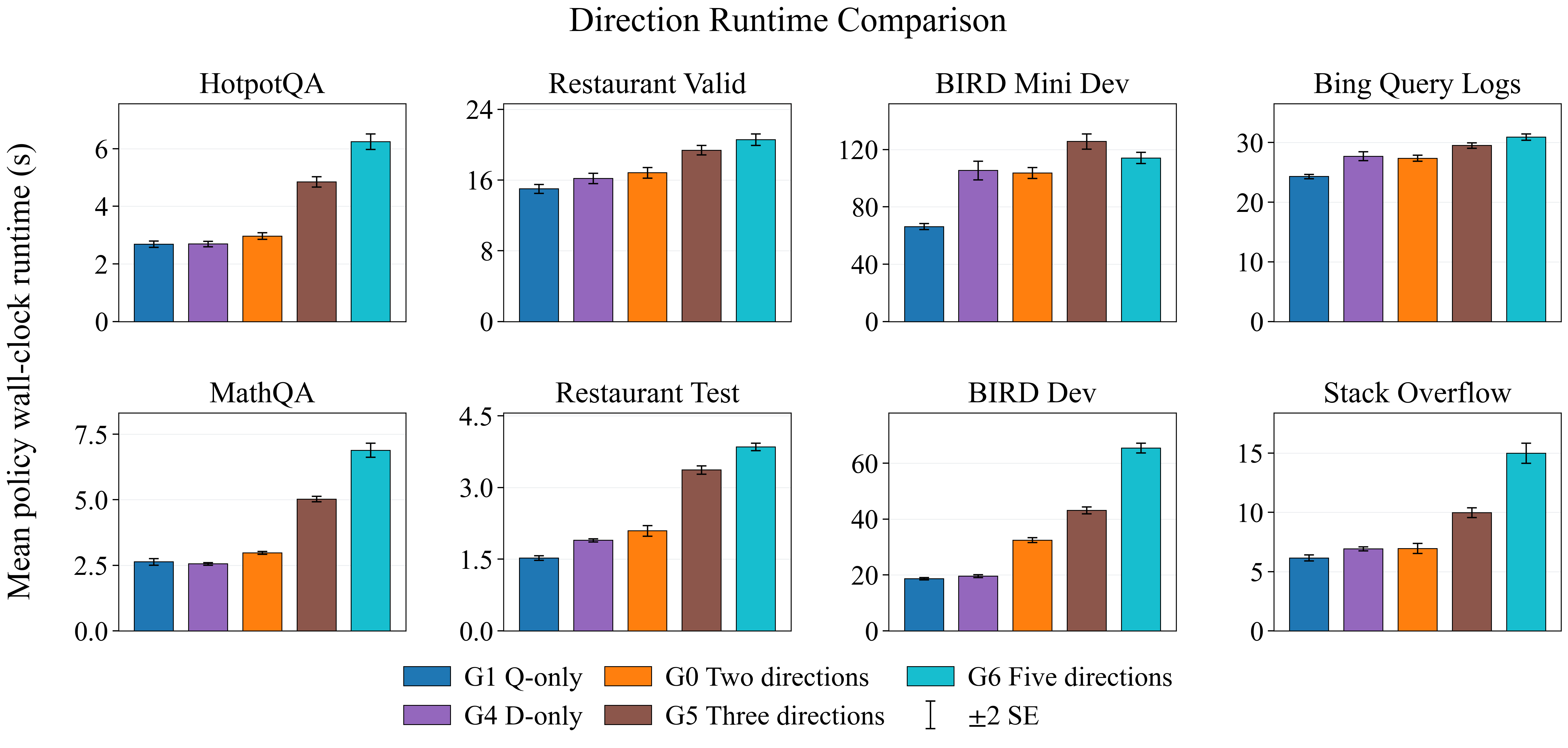}
  \caption{Direction-ablation policy runtime for a fixed workload of 30\% of matrix entries (lower is better). The five settings are Q-only G1, deployment-only G4, exact axes G0, axes plus midpoint G5, and five evenly spaced directions G6. For every dataset and matched seed, all five evaluate exactly $\lceil 0.30KN\rceil$ entries ($K$ configurations and $N$ questions), with a shortened final batch when needed; runs that reach their native stopping condition earlier continue to this common cap. They run sequentially on the same compute node, with execution order rotated across seeds. G0 has lower mean policy runtime than both denser direction sets on all eight benchmarks; G5 and G6 require 1.39$\times$ and 1.74$\times$ G0's policy time on average across benchmark ratios. Bars and error bars show means and $\pm2$ standard errors over 20 matched seeds. The timer isolates policy computation, including construction of required DP tables with reuse only within each run, and excludes data loading, result serialization, validation, and Slurm queue time.}
  \label{fig:ablation-directions-runtime}
\end{figure}

\clearpage
\subsection{Relaxation curves}

\begin{figure}[H]
  \centering
  \includegraphics[width=\textwidth,height=0.76\textheight,keepaspectratio]{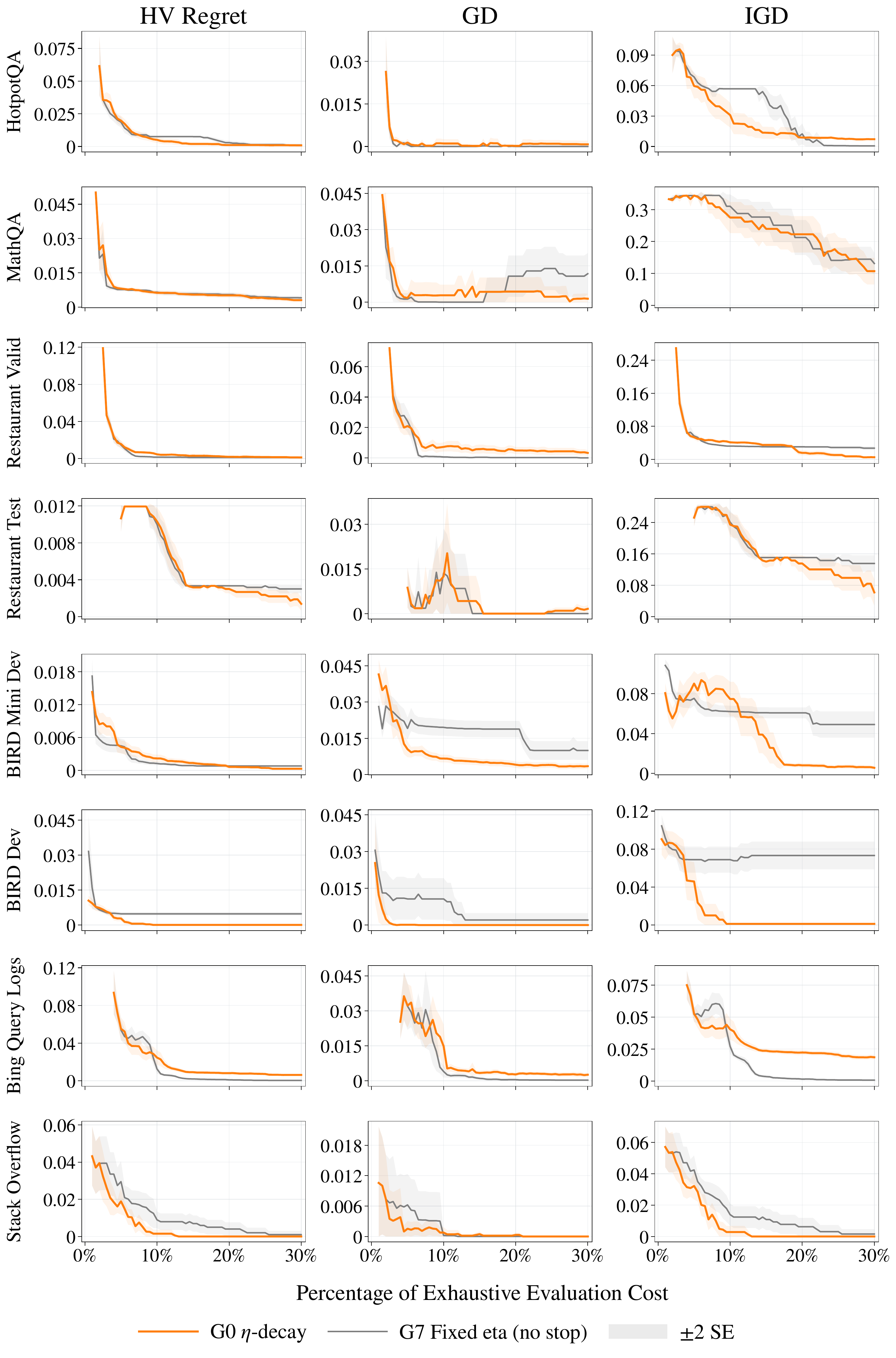}
  \caption{Relaxation ablation across eight benchmarks (lower is better). Columns report HV regret, GD, and IGD, and rows are ordered by workflow role count. G0's independent direction-local relaxation maintains competitive recommendation proximity and improves volume recovery and frontier coverage over the fixed-$\eta$ no-stop control (G7) on several benchmarks. At 30\%, G0 has lower aggregate mean HV regret, GD, and IGD, supporting our choice of the relaxation mechanism. Curves show seed means at fixed API cost checkpoints; shaded bands indicate $\pm2$ standard errors.}
  \label{fig:ablation-continuation}
\end{figure}

\clearpage
\subsection{Scheduler curves}

\begin{figure}[H]
  \centering
  \includegraphics[width=\textwidth,height=0.76\textheight,keepaspectratio]{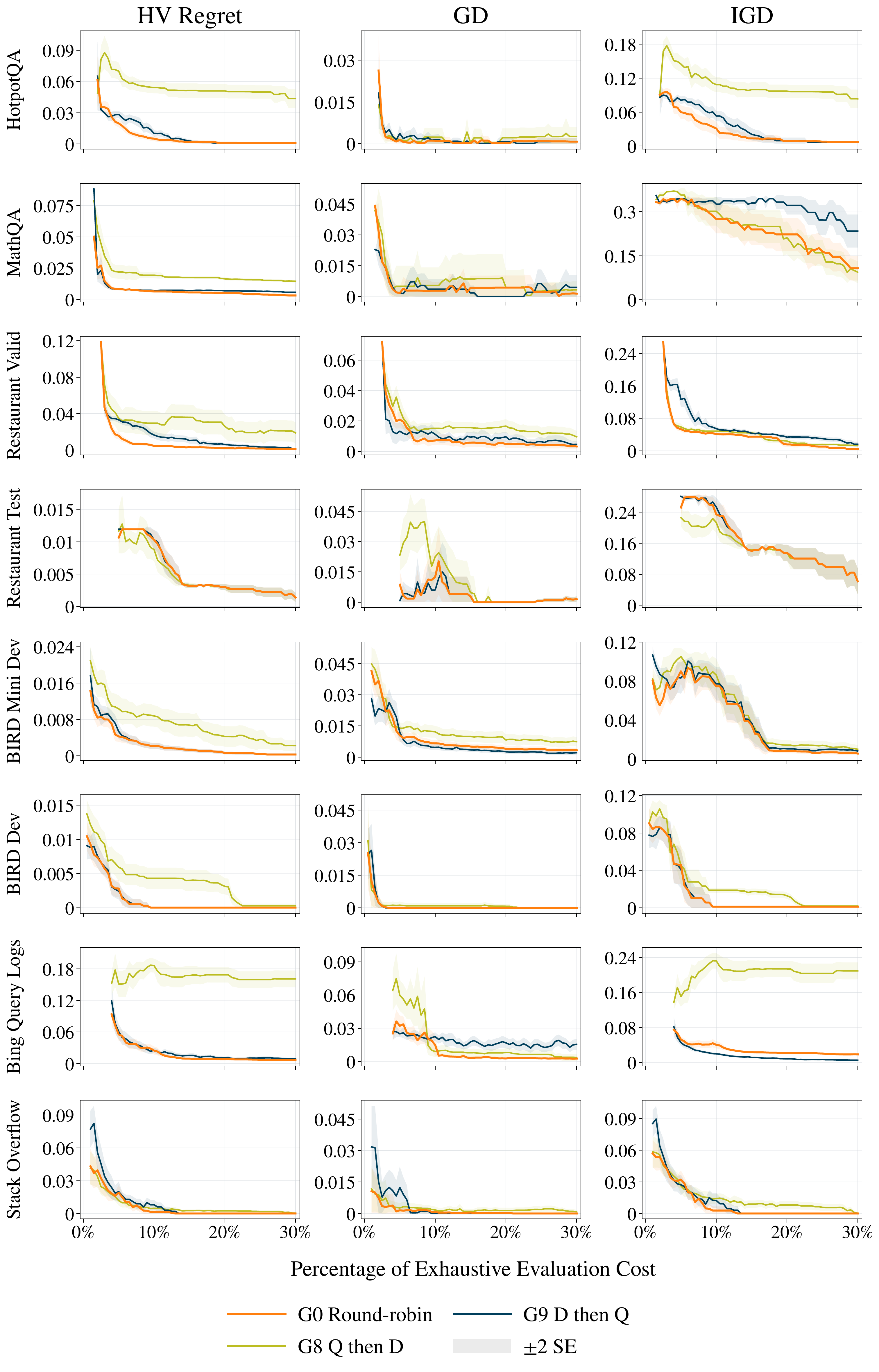}
  \caption{Scheduler ablation across eight benchmarks (lower is better). Columns report HV regret, GD, and IGD, and rows are ordered by workflow role count. G0's 1:1 round-robin interleaving achieves competitive recommendation quality across the displayed benchmarks and substantially improves on quality-first scheduling (G8) or deployment-first scheduling (G9) in several cases. At 30\%, G0 has the lowest aggregate mean of all three metrics in this family, supporting interleaving at matched cumulative search cost while acknowledging benchmark-dependent effect sizes. Curves show seed means at fixed cost checkpoints; shaded bands indicate $\pm2$ standard errors.}
  \label{fig:ablation-scheduler}
\end{figure}

\clearpage
\subsection{Prior-variance curves}

\begin{figure}[H]
  \centering
  \includegraphics[width=\textwidth,height=0.76\textheight,keepaspectratio]{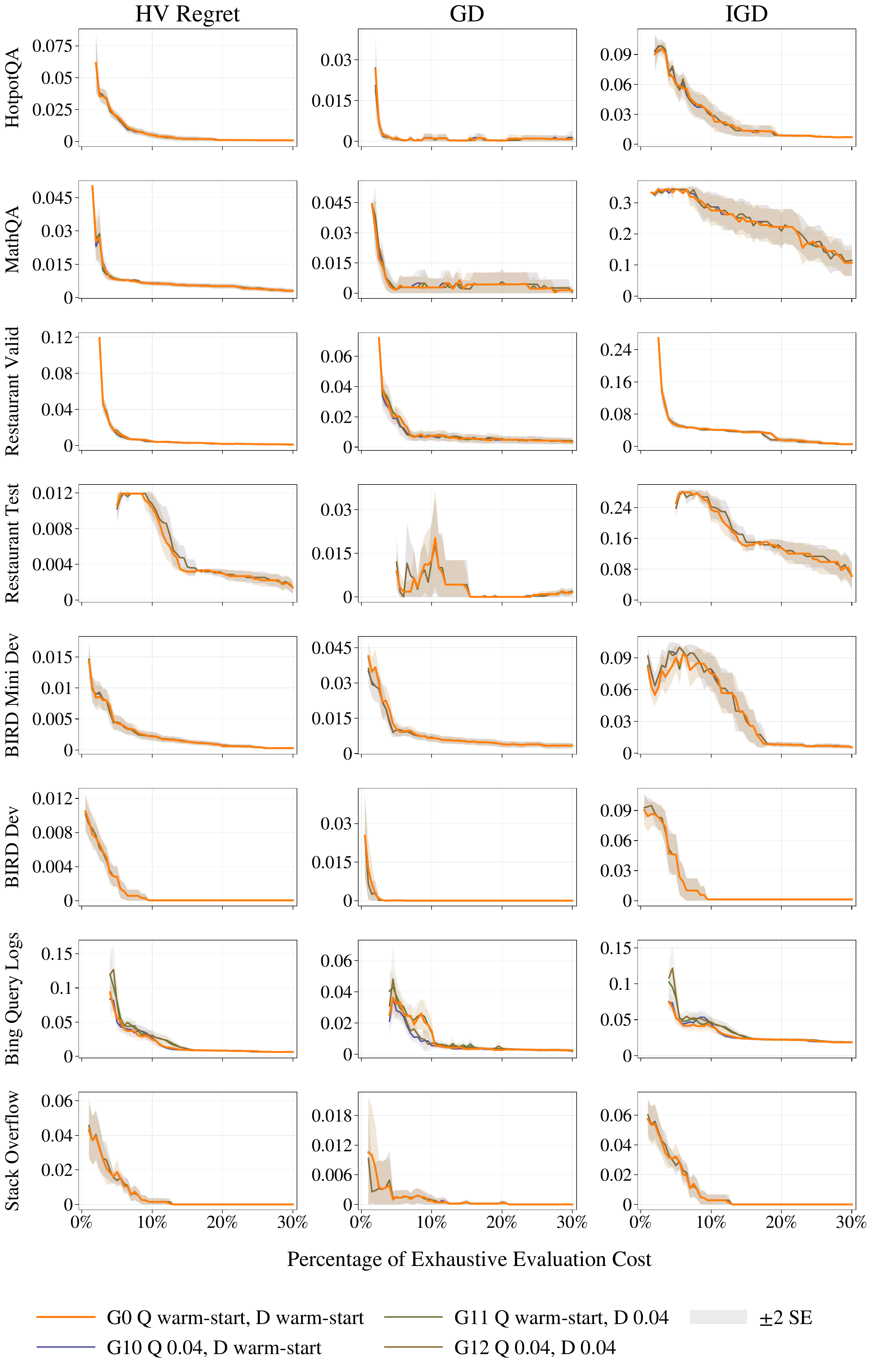}
  \caption{Prior-variance ablation across eight benchmarks (lower is better). Rows group workflows by role count; columns report HV regret, GD, and IGD. G0 estimates both quality (Q) and deployment-cost (D) prior variances from warm-start batches to calibrate initial uncertainty to the observed data. G10 fixes only Q, G11 fixes only D, and G12 fixes both. Fixed entries use variance $0.04$ in the quality or cost-desirability coordinate $C=1-D/R$; the latter corresponds to raw cost variance $0.04R^2$ in USD squared. G0 has the lowest aggregate mean GD at 30\%, while HV regret and IGD show trade-offs across settings. Prior means, observation noise, warm observations, and all other policy settings are held constant. Curves show means over 20 matched seeds; bands show $\pm2$ standard errors.}
  \label{fig:ablation-prior-variance}
\end{figure}

\clearpage
\section{Controlled Three-Objective Ablations}
\label{app:three-objective-ablations}

\paragraph{Protocol.}
We center three one-factor families and a $2\times2\times2$ prior-variance study (Fig.~\ref{fig:three-ablation-prior-variance}) on G0, the four-controller \ccgittins{} protocol with the Q, L, and D axes plus the balanced Q--L midpoint, real acquisition costs, independent direction-local $\eta$ decay, and round-robin visits. The cost family pairs Q-only real and unit pricing (G1, G7), pure-axis real and unit pricing (G4, G8), and midpoint-augmented real and unit pricing (G0, G9). The direction family compares Q-only G1, L-only G2, D-only G3, pure axes G4, G0, six Q--L-dense directions G5, and six three-edge directions G6, all with real costs and independent direction-local $\eta$ decay. The relaxation family compares G0 with fixed $\eta=1$ without Gittins stopping (G10). These identifiers refer to the three-objective controls throughout this section.

Each configuration uses the same complete HotpotQA and MathQA matrices, 20 paired question-order seeds, four warm questions per assignment, batches of up to four questions, conservative three-objective recommendation rule, and cost checkpoints spanning $0$--$30\%$ of exhaustive API evaluation cost. The frontier-metric grid contains eighteen configurations and 720 benchmark--seed runs. In the legends, Q, L, and D denote quality, latency-desirability, and deployment-cost-desirability controllers; L and D correspond to $\ell$ and $C$ in Appendix~\ref{app:extension}. At each 0.5-percentage-point target, we score each contributing seed's latest completed recommendation not exceeding that cost, retaining the final recommendation after a policy stops. Curves show means and $\pm2$ standard errors. HV regret, GD, and IGD are lower-is-better and use the fixed metric space in Appendix~\ref{app:extension}; GD and IGD score the complete returned set.

\paragraph{Cost mechanism.}
Figure~\ref{fig:three-ablation-cost} separates acquisition pricing from controller choice. G1 versus G7, G4 versus G8, and G0 versus G9 compare real with unit prices while retaining observed deployment costs, the Pareto recommendation rule, and realized-dollar accounting. For the four-controller policy, using observed acquisition costs lowers HV regret and GD on both matrices at 30\%. This result supports the real-cost pricing used by G0 to allocate its evaluation budget.

\paragraph{Directions.}
Figure~\ref{fig:three-ablation-directions} compares how controller directions allocate evaluation effort. All controllers observe the same three outcomes, which a common recommendation rule uses to form the returned set. G4 interleaves the three pure axes; G0 adds the Q--L midpoint with radial utility $\min\{Q_i,\ell_i\}$. G5 combines the Q--L endpoints, quarter points, and midpoint with the D axis; G6 combines the three axes with the Q--L, L--D, and Q--D midpoints. Round-robin scheduling balances decision opportunities, not purchased batches or dollars. At 30\%, G0 has lower HV regret than each single-direction control on both matrices. Its four directions represent all three objective axes and the quality--latency interaction with fewer controllers than the six-direction variants, providing a compact main configuration.

Figure~\ref{fig:three-ablation-directions-runtime} compares policy computation at an equal number of evaluated entries. G0 takes 13.3 seconds on HotpotQA and 12.5 seconds on MathQA, within the 10.0--15.0-second range of the tested direction sets. Together with its low HV regret at 30\% of evaluation cost (Figure~\ref{fig:three-ablation-directions}), this runtime supports using four directions as the main three-objective configuration.

\paragraph{Relaxation.}
Figure~\ref{fig:three-ablation-continuation} isolates the continuation rule under the same four controllers and round-robin schedule. G0 independently relaxes each direction's evaluation-cost penalty after local stopping signals, whereas G10 holds $\eta=1$ fixed. This direction-local adaptation retains the stopping behavior used by the main anytime policy; the curves show how the alternative continuation rules affect the three frontier metrics on each benchmark.

\clearpage
\subsection{Cost-mechanism curves}

\begin{figure}[H]
  \centering
  \includegraphics[width=\textwidth]{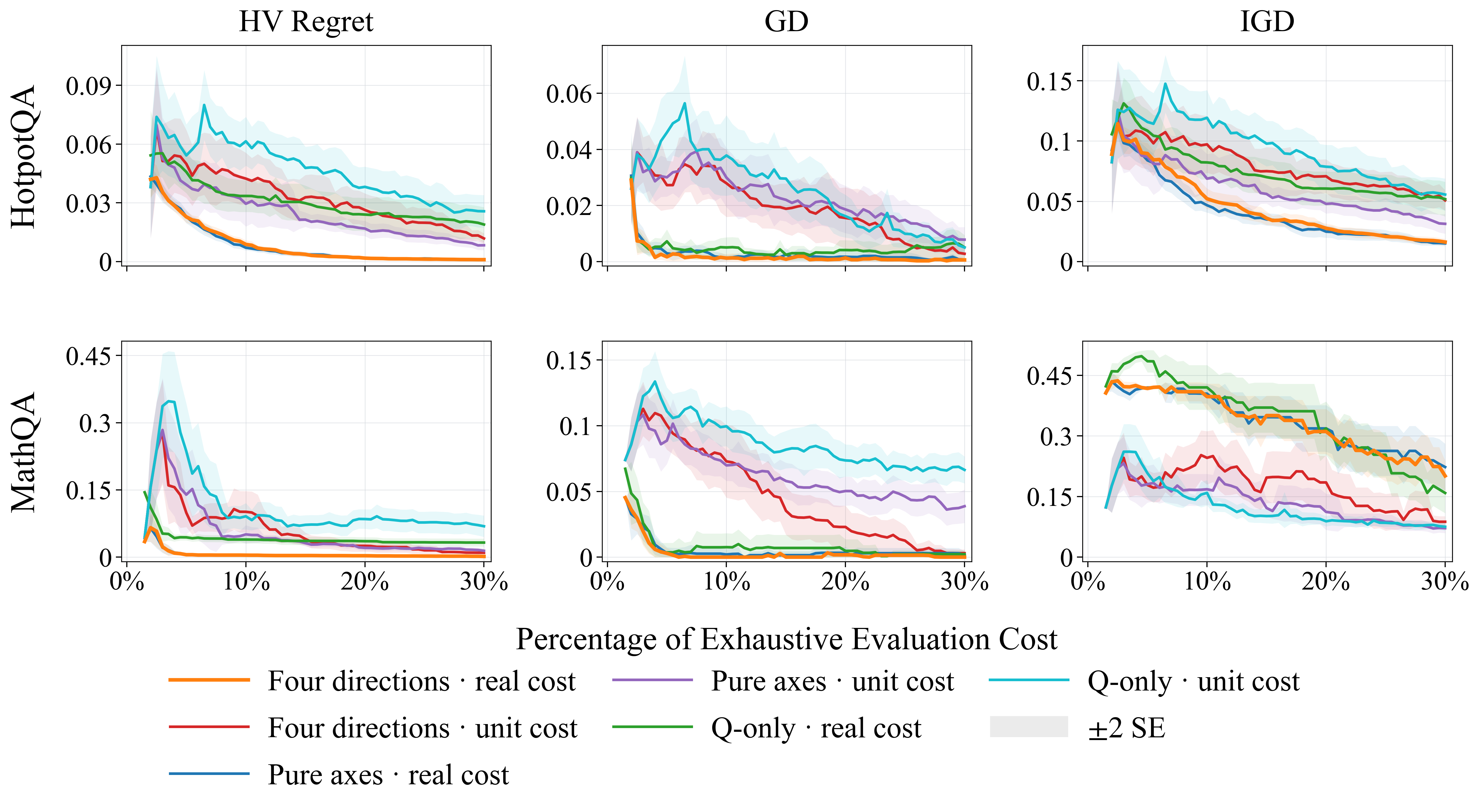}
  \caption{Three-objective cost-mechanism ablation on HotpotQA (top) and MathQA (bottom). Columns report HV regret, GD, and IGD (lower is better). G0 combines real acquisition costs with four controllers and achieves lower mean HV regret and GD than its unit-cost counterpart G9 on both benchmarks at 30\%. The comparison isolates the effect of acquisition pricing on the four-controller policy. The Q-only pair G1/G7 and pure-axis pair G4/G8 provide corresponding pricing controls. These comparisons vary only the acquisition penalty, retaining observed deployment costs, the Pareto recommendation rule, and realized-dollar accounting. Curves show seed means and bands show $\pm2$ standard errors.}
  \label{fig:three-ablation-cost}
\end{figure}

\clearpage
\subsection{Direction curves}

\begin{figure}[H]
  \centering
  \includegraphics[width=\textwidth]{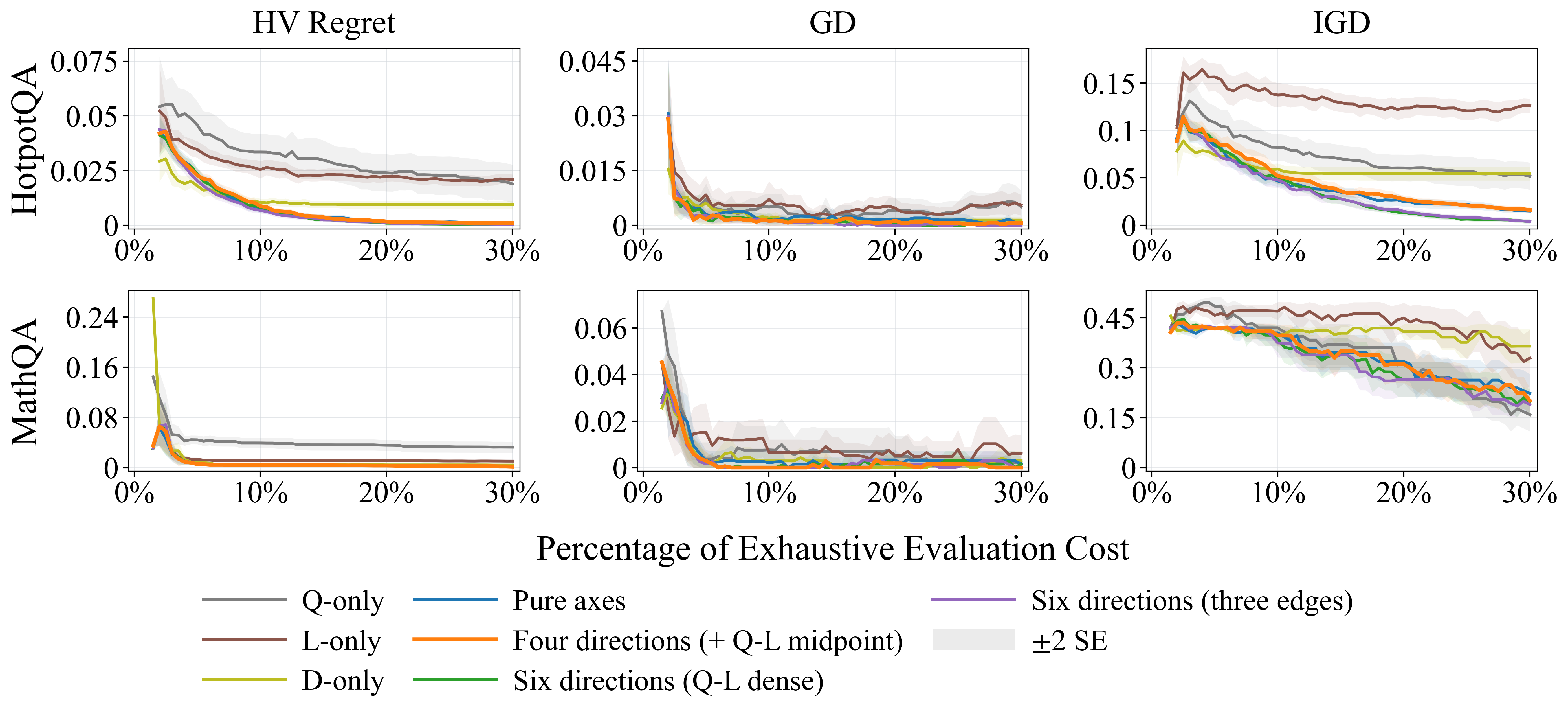}
  \caption{Three-objective direction ablation on HotpotQA (top) and MathQA (bottom). Columns report HV regret, GD, and IGD (lower is better). G0 combines the three pure axes with the Q--L midpoint, achieving strong volume recovery and low GD on both benchmarks. At 30\%, its mean HV regret is lower than that of all three single-direction controls G1, G2, and G3. Increasing the direction count does not uniformly improve all three frontier metrics, and coverage remains benchmark- and budget-dependent. All variants use real acquisition costs; curves show seed means and bands show $\pm2$ standard errors.}
  \label{fig:three-ablation-directions}
\end{figure}

\begin{figure}[p]
  \centering
  \includegraphics[width=\textwidth]{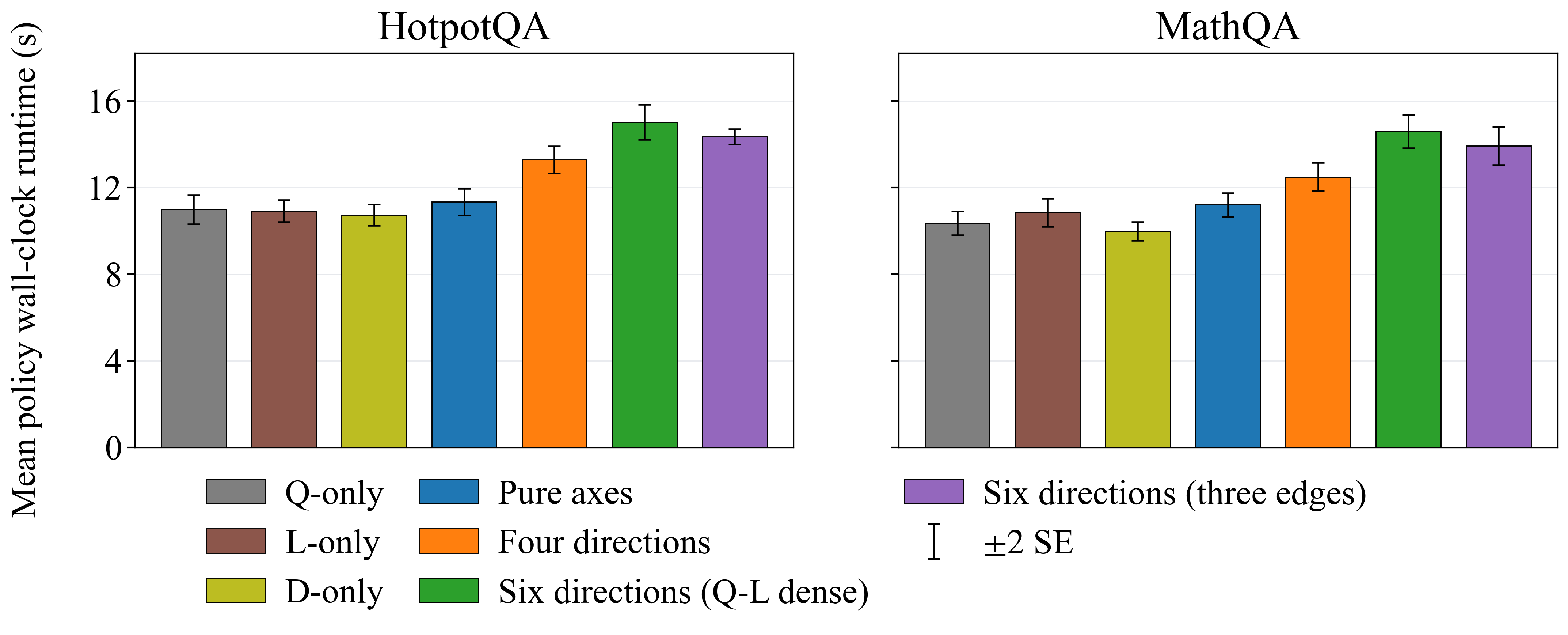}
  \caption{Three-objective direction-ablation policy runtime for a fixed workload of 30\% of matrix entries (lower is better). G0's four-direction configuration takes 13.3 seconds on HotpotQA and 12.5 seconds on MathQA, within the 10.0--15.0-second range across all tested variants. The four-controller policy remains within the runtime range of the tested direction sets. For every benchmark and matched seed, all seven variants evaluate exactly $\lceil0.30KN\rceil=6{,}000$ entries ($K=100$ configurations and $N=200$ questions), with a shortened final batch when needed; native stopping signals do not terminate these runs before the common cap. Bars and error bars show means and $\pm2$ standard errors over 20 matched seeds. The timer isolates online policy computation and excludes data loading, offline oracle analysis, result serialization, and Slurm queue time. Runs use the same code and data and are independently scheduled on HPC compute nodes, so node conditions may differ.}
  \label{fig:three-ablation-directions-runtime}
\end{figure}

\clearpage
\subsection{Relaxation curves}

\begin{figure}[H]
  \centering
  \includegraphics[width=\textwidth]{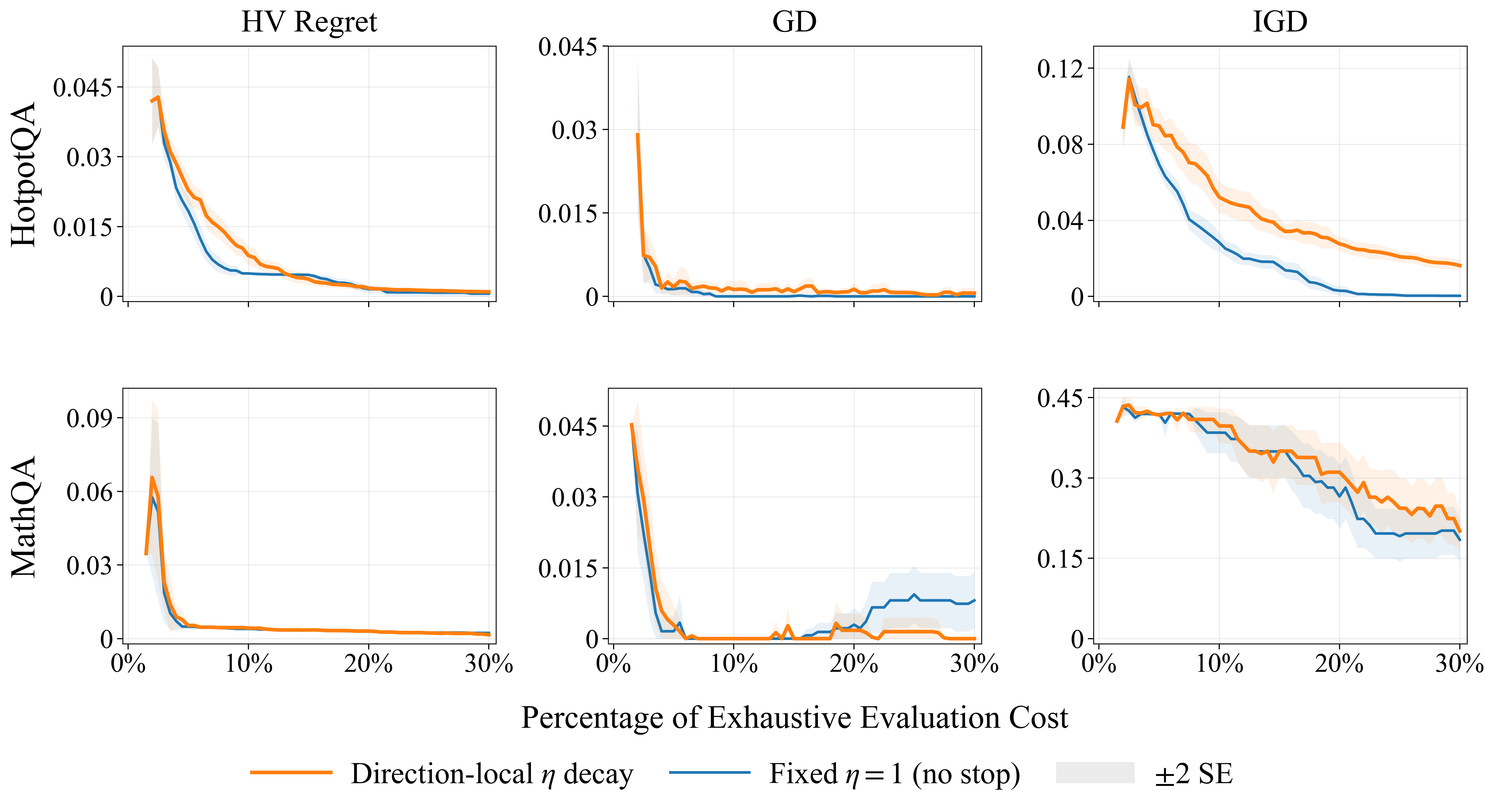}
  \caption{Three-objective relaxation ablation on HotpotQA (top) and MathQA (bottom). Columns report HV regret, GD, and IGD (lower is better). G0 applies independent direction-local relaxation to four real-cost controllers: each multiplier starts at $\eta=1$ and is halved after a local stopping signal when further relaxation is allowed. G10 holds $\eta=1$ fixed without local Gittins stopping. The curves compare the resulting volume recovery, recommendation proximity, and frontier coverage under the same evaluation protocol. Both policies use equal-frequency round-robin visits; curves show seed means and bands show $\pm2$ standard errors.}
  \label{fig:three-ablation-continuation}
\end{figure}

\clearpage
\subsection{Prior-variance curves}

\begin{figure}[H]
  \centering
  \includegraphics[width=\textwidth]{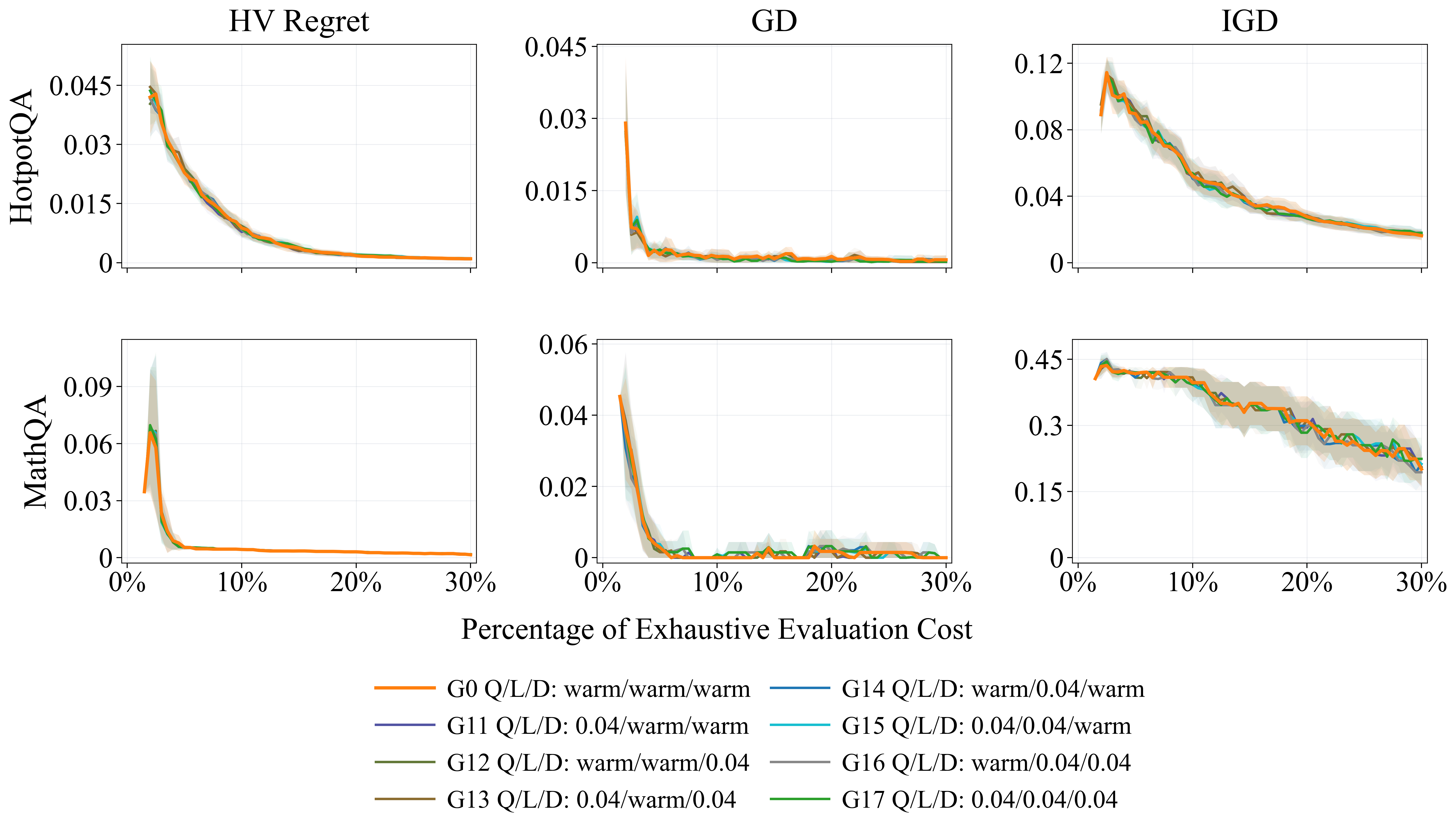}
  \caption{Three-objective prior-variance ablation on HotpotQA (top) and MathQA (bottom). Columns report HV regret, GD, and IGD (lower is better). The $2\times2\times2$ design crosses warm-start estimation and fixed variance for quality (Q), latency desirability (L), and deployment-cost desirability (D). G0 estimates all three variances from warm-start batches; G11--G17 cover the other seven combinations, as specified in the legend. Fixed entries use variance $0.04$ in the corresponding desirability coordinate, equivalent to raw latency variance $0.04T^2$ in seconds squared and raw cost variance $0.04R^2$ in USD squared. Prior means, observation noise, warm observations, and all other policy components are held constant. At 30\%, G0 has low HotpotQA GD and zero MathQA GD. HV regret and IGD show benchmark-dependent trade-offs across prior settings. The main policy uses warm-start estimates for all three priors to calibrate initial uncertainty from the observed data. Curves show means over 20 matched seeds; bands show $\pm2$ standard errors.}
  \label{fig:three-ablation-prior-variance}
\end{figure}

\end{document}